\documentclass{article} 
\usepackage{iclr2026_conference,times}
\usepackage{graphicx}
\usepackage{amsmath,amssymb}     

\usepackage{amsmath,amsfonts,bm}

\def\eqref#1{equation~\ref{#1}}

\def\1{\bm{1}}

\DeclareMathAlphabet{\mathsfit}{\encodingdefault}{\sfdefault}{m}{sl}
\SetMathAlphabet{\mathsfit}{bold}{\encodingdefault}{\sfdefault}{bx}{n}

\makeatletter
\renewcommand{\@oddhead}{}
\renewcommand{\@evenhead}{}
\makeatother
\usepackage{hyperref}
\usepackage{url}
\usepackage{booktabs}
\usepackage{multirow}
\usepackage[dvipsnames,table]{xcolor} 
\usepackage{colortbl}   
\usepackage{graphicx}
\usepackage{float}
\usepackage{CJK}
\usepackage{enumitem}

\renewcommand{\cite}[1]{\citep{#1}}

\title{SpaCE-10: A Comprehensive Benchmark for Multimodal Large
Language Models in Compositional Spatial Intelligence }

\iclrfinalcopy
\author{Ziyang Gong$^{1,3}$\thanks{Equal contribution},
Wenhao Li$^{2,3*}$,
Oliver Ma$^{3*}$, Songyuan Li$^{4}$, Zhaokai Wang$^{1,3}$,\\\textbf{Songze Li}$^{3,5}$,\quad \textbf{Jiayi Ji}$^{2,6}$,\quad \textbf{Xue Yang}$^{1}$,\quad \textbf{Gen Luo}$^{3}$,\quad \textbf{Junchi Yan}$^{1}$,\quad \textbf{Rongrong Ji}$^{2}$  \\
$^{1}$SJTU\quad $^{2}$XMU \quad $^{3}$Shanghai AI Lab \quad $^{4}$SYSU \quad $^{5}$ FDU \quad $^{6}$NUS \\{\tt\small\url{https://github.com/VisionXLab/SpaCE-10}}
}

\begin{document}







%

\maketitle

\begin{abstract}
Multimodal Large Language Models (MLLMs) have achieved remarkable progress in various multimodal tasks. To pursue higher intelligence in space, MLLMs require integrating multiple spatial capabilities, even for handling simple and normal tasks.  However, existing benchmarks struggle to comprehensively evaluate the spatial intelligence of common MLLMs from the atomic level to the compositional level. To fill this gap, we present SpaCE-10, a comprehensive benchmark for compositional spatial evaluations. In SpaCE-10, we define 10 atomic spatial capabilities, which are combined to form 8 compositional capabilities. Based on these definitions, we propose a novel hierarchical annotation pipeline to generate high-quality and diverse question-answer (QA) pairs.  With over 150+ hours of human expert effort, we obtain over 5\textit{k} QA pairs for 811 real indoor scenes in SpaCE-10, which covers various evaluation settings like point cloud input and multi-choice QA.  We conduct an extensive evaluation of common MLLMs on SpaCE-10 and find that even the most advanced MLLM still lags behind humans by large margins. Through our careful study, we also draw several significant findings that benefit the MLLM community. For example, we reveal that the shortcoming of counting capability greatly limits the compositional spatial capabilities of existing MLLMs. 
\end{abstract}

\section{Introduction}

\begin{figure}[t]
    \includegraphics[width=0.98\linewidth]{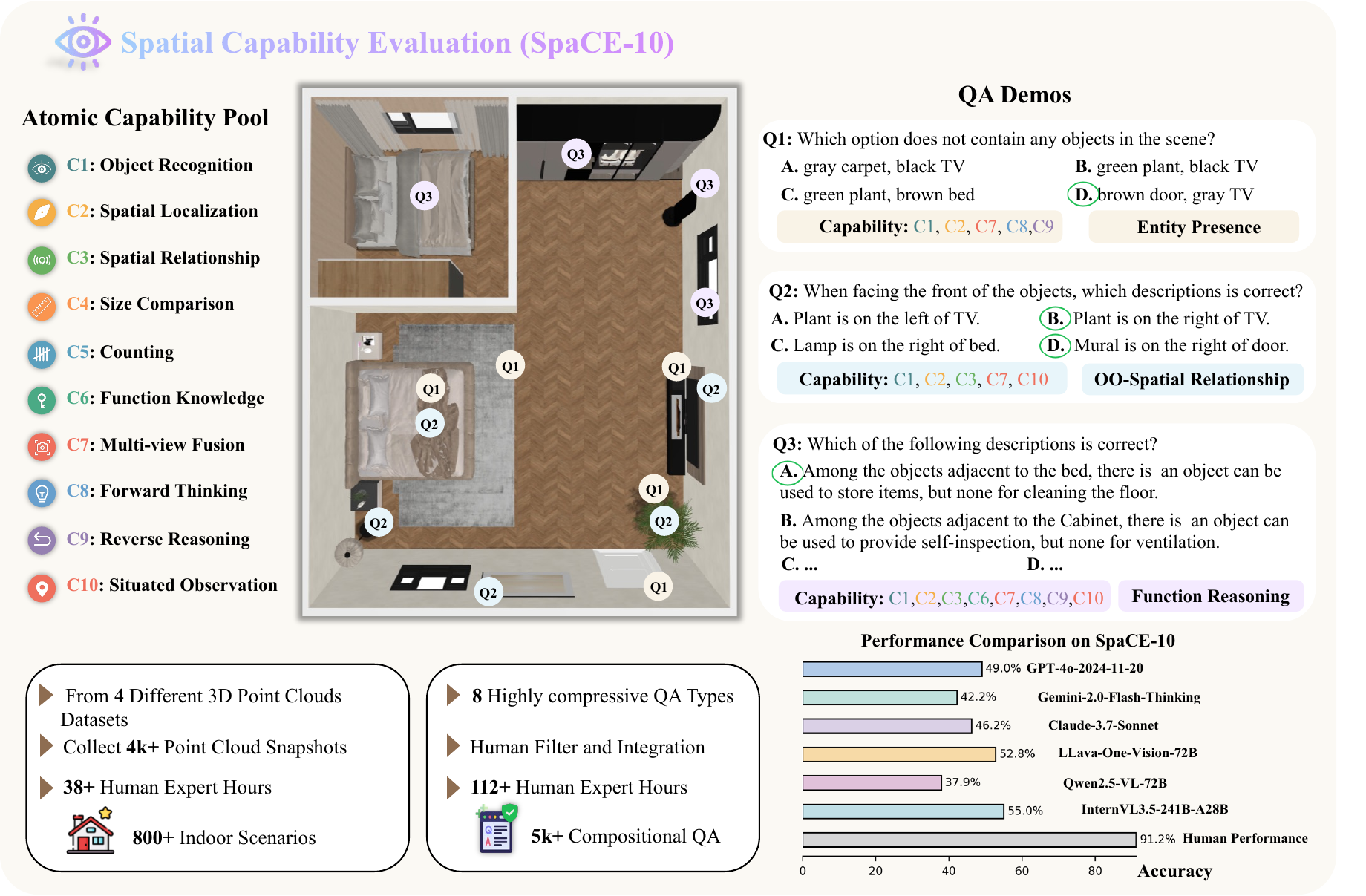}
    \vspace{-0.3cm}
    \caption{\textbf{Overview of SpaCE-10 benchmark.} SpaCE-10 takes over 150 human expert hours to collect 5\textit{k}+ QA pairs in  811 indoor scenes, which can evaluate MLLMs from 10 atomic capabilities to 8 compositional capabilities.   Through evaluations, SpaCE-10 indicates that even the most advanced MLLM  still lags far behind humans by large margins. Green cirle means the correct answer.} 
    \label{fig: teaser}
    \vspace{-0.4cm}
\end{figure}

Recent years have witnessed the rapid development of multimodal large language models (MLLMs)~\cite{GPT-5,qwen2.5-vl,internvl3.5}, which continually narrows the gap between machines and humans in multimodal tasks~\cite{mmbench}. The significant progress has motivated researchers to pursue higher machine intelligence in the real world~\cite{leo,3d-llm,vebrain}. Focus on the scene in Fig.~\ref{fig: teaser},
imagine you are about to head out the door and tell your home robot, \textit{`I forgot my watch, please bring it to me. I remember it's near the nightstand.'} To succeed, the robot must know what a watch is, plan to the nightstand, reason about spatial relations near, localize the watch among distractors, retrieve it, and return under diverse changing viewpoints. Solving this `simple, everyday task' requires the on-the-fly composition of a diverse set of spatial capabilities. This raises a central question: \textit{Can current MLLMs master these spatial capabilities and compose them seamlessly in real-world scenarios?}

While existing benchmarks have made valuable explorations of the spatial intelligence of multimodal large language models (MLLMs)~\cite{vsi,3dsrbench,msqa,mmsi,omnispatial}, they seldom make these capabilities explicit or design tasks that systematically combine them. 
Early benchmarks~\cite{scanqa,sqa3d,clever3d,3dqa,scanrefer} mainly focus on the assessment of less-combined capabilities like object recognition and spatial localization, while compositional ones still remain to be defined and evaluated.  Recent benchmarks~\cite{vsi,mmsi,omnispatial} aim to evaluate the spatial intelligence of MLLMs through more compositional questions, but still fail to reflect the role of different spatial capabilities in compositional reasoning.  More importantly,  existing spatial benchmarks struggle to satisfy the evaluation needs of existing MLLMs in terms of scenes,  modalities, and question types, \textit{etc}. As shown in Tab.~\ref{table: dataset_comparison},  the number of scenarios in existing benchmarks is usually less than 400, which makes it difficult to cover various practical situations.

To fill these gaps, this paper proposes SpaCE-10 (Spatial Capability Evaluation), a capability-focus question-answer (QA) benchmark with an atomic capability pool (Fig.~\ref{fig: teaser}). This pool highlights the 10 core spatial capabilities 
(C1-C10) for MLLMs in real-world deployment. 
In SpaCE-10, there are 8 meticulously designed and systematically combined QA types, each of which covers more than 5 atomic capabilities. Hence, SpaCE-10 can not only assess the Compositional Spatial Intelligence (CSI) of MLLMs but can also reflect the impact of different atomic capabilities in spatial comprehension.

Based on this design principle, we propose an innovative hierarchical annotation pipeline in SpaCE-10. Specifically,  we collect over 800 real indoor scanned scenes from four public datasets.  For each scene, we present an automated pipeline to generate structured data that can describe different types of information in the scene, \emph{e.g.,} appearance and relationship.  Based on this information, a multi-stage semi-automated pipeline is adopted to generate basic QA pairs, conduct quality verification, and perform the capability integration.   Our SpaCE-10 consists of more than 5,000 high-quality QA pairs,  covering various settings of existing MLLMs, \emph{e.g.,}  point cloud input and multi-choice question types.  As shown in Tab.~\ref{table: dataset_comparison}, SpaCE-10 demonstrates greater diversity than previous benchmarks in data distribution, annotation process, and evaluation settings, showing promising all-around evaluation ability for compositional spatial intelligence. 
  
We conduct extensive and systematic evaluations of mainstream MLLMs on  SpaCE-10,  including 4 close-source MLLMs and nearly \textbf{50} open-source MLLMs ranging from 1B to 241B. Experimental results show that even the most advanced MLLMs are still far behind humans in compositional spatial intelligence, \emph{i.e.,} 53.1\% of GPT-5 \textit{vs.} 91.2\% of human. Meanwhile, 2D MLLMs demonstrate much stronger capabilities than 3D MLLMs on SpaCE-10, showing great potential for image-based spatial reasoning.  In addition, existing MLLMs greatly fall short in multiple-answer QAs, suggesting their inferior complex reasoning abilities. Our further study also reveals that the shortcoming of counting capability greatly limits the compositional spatial capabilities of existing MLLMs. These findings provide valuable directions for the community to develop more capable MLLMs in terms of spatial intelligence.  Overall, our main contributions are summarized as follows:

\begin{itemize}[leftmargin=*,itemsep=1pt, topsep=0pt, parsep=0pt]
    \item We present SpaCE-10, a comprehensive benchmark for compositional spatial intelligence. SpaCE-10 is the most diverse benchmark that can assess the capabilities of MLLMs from the atomic level to the compositional level.  It also covers various evaluation settings, including 3D inputs and multi-choice questions.

    \item  We propose an innovative hierarchical annotation pipeline in SpaCE-10, which first produces structured descriptions of scenes via an automated pipeline and then generates compositional QA pairs through a multi-stage semi-automated pipeline. The hierarchical pipeline ensures the quality, diversity, and controllability of the generated QA pairs.

    \item  We conduct extensive evaluations for nearly 50 open- and close-source MLLMs on  SpaCE-10
    Through our in-depth analysis, we draw several significant findings that will benefit the spatial intelligence of future MLLMs in the community. 
    
\end{itemize}

\section{Related Work}

\begin{table}[!t]
\scriptsize
\caption{\textbf{Comparison of SpaCE-10 with existing spatial benchmarks.} Our SpaCE-10 contains the most diverse scenarios and QA types, covering various evaluation settings of existing MLLMs. CSI means the investigation of Compositional Spatial Intelligence. SCN, OBJ, HM3D, 3RS, ARK represent ScanNet~\cite{scannet}, Objaverse, Habitat-Matterport-3D~\cite{hm3d}, 3RScan~\cite{3rscan}, and ArkitScene~\cite{arkitscenes}. Sim. denotes similarity-based metrics. 2D\&3D means the benchmarks support both 2D imagery and 3D point cloud-based MLLMs.
}
\vspace{3mm}
\renewcommand{\arraystretch}{1.2}
  \resizebox{\linewidth}{!}{
\begin{tabular}{@{}lccccccc@{}}
\toprule
\textbf{Dataset}  & \textbf{Scenario Source }          & \textbf{Scene} & \textbf{Q\&A} &\textbf{Metric}   & \textbf{2D \& 3D }& \textbf{Multi-Answer }& \textbf{CSI} \\ \midrule
3DQA~\cite{fe-3dgqa}     & SCN               & -      & 902        & Sim.     &      $\times$               &     $\times$   &   $\times$    \\
ScanQA~\cite{scanqa}   & SCN                & 167    & 10k        & Sim.     &         $\checkmark$    &    $\times$    &  $\times$    \\
FE-3DGQA~\cite{fe-3dgqa} & SCN                  & 100    & 3.9k       & Sim.     &        $\times$     &      $\times$  & $\times$ \\
SQA3D~\cite{sqa3d}    & SCN                  & 132    & 6.9k       & Sim.     &    $\checkmark$      &      $\times$  &   $\times$   \\
CLEVER3D~\cite{clever3d} & SCN                  & 133    & 10k        & Sim.     &      $\times$    &   $\times$   &    $\times$  \\
3D-LLM~\cite{3d-llm}   & OBJ,SCN,HM3D      & -      & 30k        & Sim.     &    $\times$     &   $\times$ &  $\times$   \\
M3DBench~\cite{m3dbench} & SCN               & -      & 1.5k       & Sim.+LLM &   $\times$       &   $\times$  &    $\times$ \\
MSQA~\cite{msqa}     & SCN,3RS,ARK        & 381    & 3.5k       & LLM      &     $\times$   &    $\times$   &   $\times$  \\
VSI~\cite{vsi}      & SCN,3RS,SCN++      & 288    & 5.0k       & Acc.     &   $\times$    &    $\times$  & $\times$\\
\rowcolor{gray!15} \textbf{SpaCE-10 (Ours)}    & \textbf{SCN, 3RS, ARK, SCN++}  & \textbf{811}    & \textbf{5.0k}         & \textbf{Acc.}     &  $\checkmark$     &   $\checkmark$    &  $\checkmark$  \\ \bottomrule
\end{tabular}}
\vspace{-0.6cm}
\label{table: dataset_comparison}
    \end{table}
    
Early works start spatial intelligence mainly with two directions: (i) 2D abstract reasoning with logic-puzzle and geometric-panel tasks (e.g., Raven-style matrices)~\cite{raven,logicvista,visulogic,does}, and (ii) simplified images with only a few objects, where queries test basic relations such as above/below/left/right/size~\cite{eye_wide_shut,whatsup,elephants}. As attention shifts to realistic environments, 3D scene benchmarks~\cite{scanqa,sqa3d,msqa,3dqa,m3dbench} emerge and expand to richer tasks, such as route planning and situated perception from specified viewpoints. However, they typically adopt point-cloud inputs and still treat spatial ability as a single block. With the rise of MLLMs, newer studies~\cite{3dsrbench,vsi,mmsi,omnispatial,egoexobench} examine 3D spatial comprehension directly from 2D images or videos. Yet most works do not make capabilities explicit or delve into them. To this end, SpaCE-10 defines an atomic capability pool (C1-C10) across perception and reasoning to assess a model's CSI. It also supports both 2D images and 3D point clouds, offering a new perspective for advancing spatial intelligence in MLLMs.

\section{SpaCE-10}
\subsection{Overview}
\textbf{Construction} The images in SpaCE-10 are from four 3D indoor-scene scan datasets, including ScanNet++ , ScanNet (SCN), 3RScan (3RS), and ARkitScene (ARK) with over 800 real indoor scenes, including a wide variety of environments such as living rooms, classrooms, bathrooms, kitchens, and more. Finally, SpaCE-10 consists of 8 QA types that are EQ (Entity Quantification), SQ (Scene Quantification), SA (Size Assessment), OO (Object-Object Spatial Relationship), OS (Object-Scene Spatial Relationship), EP (Entity Presence), FR (Functional Reasoning), and SP (Spatial Planning). Also it includes 10 atomic spatial capabilities of C1 (Object Recognition), C2 (Spatial Localization), C3 (Spatial Relationship), C4 (Size Comparison), C5 (Counting), C6 (Funciton Knowledge), C7 (Multi-view Fusion), C8 (Forward Thinking), C9 (Reverse Reasoning), and C10 (Situated Observation). Definition of QA and capabilities are in Sec.~\ref{sec:qa_definition} and \ref{sec:cap_definition}, respectively.

\begin{figure}[!t]
    \centering
    \includegraphics[width=0.95\linewidth]{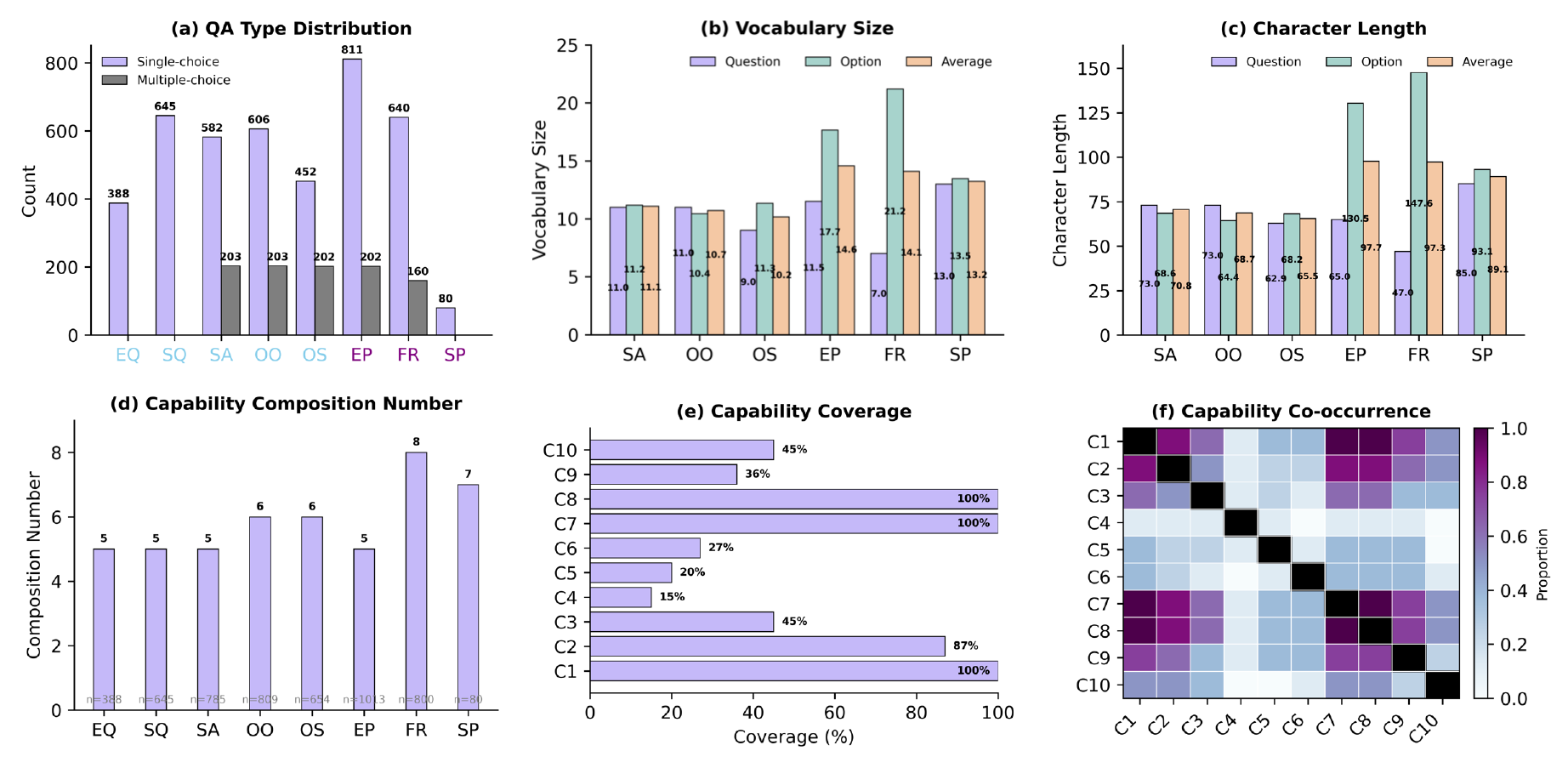}
        \vspace{-0.4cm}
    \caption{\textbf{Dataset analysis of SpaCE-10.} (a) Number distribution of each QA type. SpaCE-10 consists of 8 QA types that are EQ (Entity Quantification), SQ (Scene Quantification), SA (Size Assessment), OO (Object-Object Spatial Relationship), OS (Object-Scene Spatial Relationship), EP (Entity Presence), FR (Functional Reasoning), and SP (Spatial Planning).  (b) Average vocabulary size per QA type for question, option, and average. (c) Average character length per QA type. (d) Coverage of the atomic capabilities (C1-C10). (e) The correlation between human expert accuracy and average character length across six QA types. (f) Capability co-occurrence matrix.}
\vspace{-0.8cm}
    \label{fig:statistic}
\end{figure}

\textbf{Analysis} In Fig.~\ref{fig:statistic} (a), we show the numbers of each type of QA in SpaCE-10. Among them, blue ones belong to Perception, and purple ones belong to Reasoning. In (b) and (C), we demonstrate the average vocabulary size (unique word number) and character length of the Question and Option of 6 QA types. EQ and SQ are excluded because their options are numbers.


For Capability, in (d) we show the number of capability contained in each QA type. (e) represents the capability coverage in each
QA type, revealing a three-tier hierarchy. (1) \textbf{Foundation}: C1, C7, C8 appear in 100\% of QA types since they are prerequisites for almost every task. (2) \textbf{Bridge}: C2 covers 87\% and links both `quantification' and `relation/plan' QA types. (3) \textbf{Specialist}: C3 and C10 are mid-frequency (45\%), activated when viewpoint-dependent relations matter; C9 (36\%) attaches to causal/plan problems; C4 (15\%), C5 (20\%), and C6 (27\%) appear in specific QA types. 
(f) shows the co-occurrence of each capability. Beyond of (d), we observe the tightest pair is C3 with C10 = 4/8 (OO, OS, FR, SP), which means they appear together in 3 QA types, showing that when relative relations are asked, a specified viewpoint is usually required. Moreover, C9 appears mainly with planning-style tasks: C2 with C9 = 3/8 (EP, FR, SP), while C3 with C9 = 2/8 (FR, SP) and C10 with C9 = 2/8 (FR, SP). The design motivation of the atomic capability pool can be found at Sec.~\ref{sec:capability}.

\vspace{-0.1cm}

\subsection{Hierarchical Annotation Pipeline}

\begin{figure}[!t]
    \includegraphics[width=0.98\linewidth]{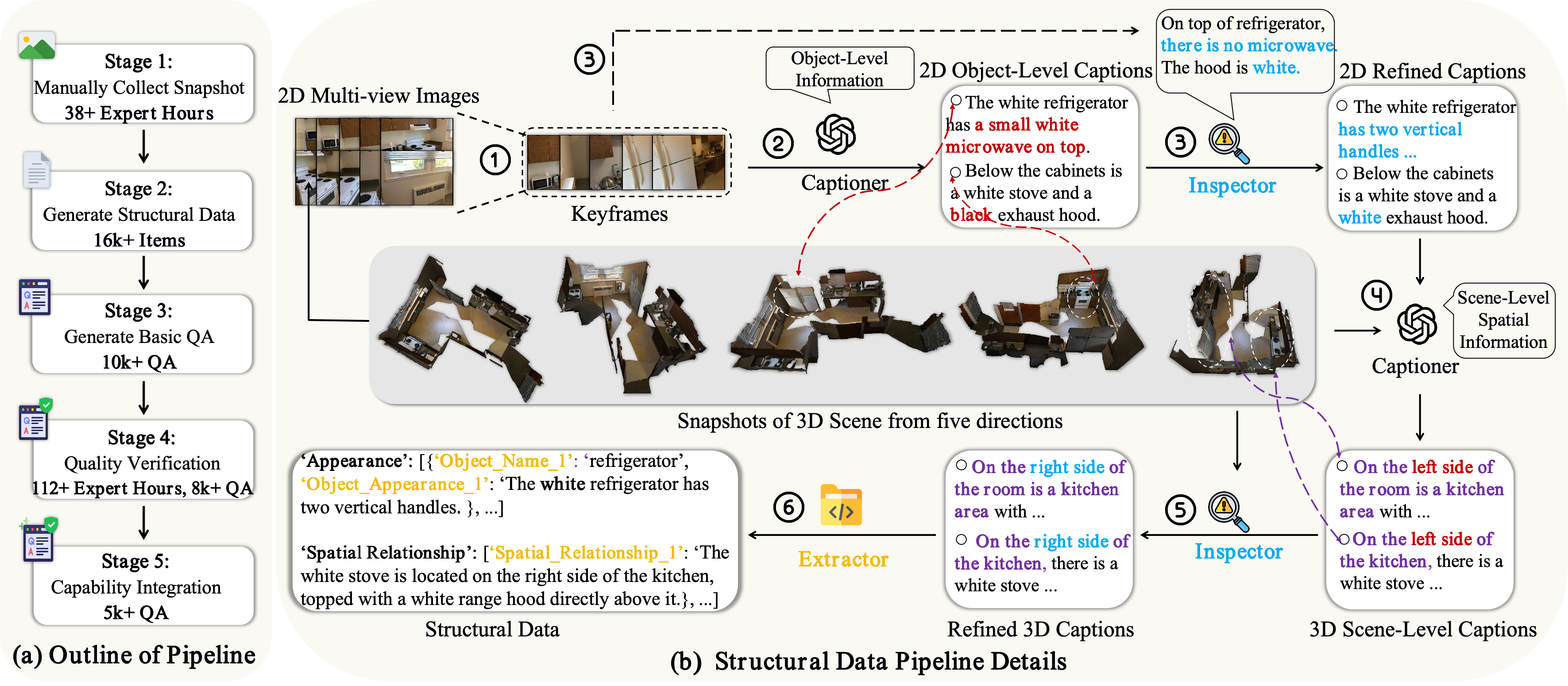}
        \vspace{-0.2cm}
    \caption{\textbf{Illustration of our hierarchical annotation pipeline.} We  generate structural data to construct over 10\textit{k} QA pairs, and performs capability integration to obtain over 5\textit{k} QA pairs with 10 compositional capabilities. This process takes over 150 expert hours for data collection and filtering.} 
    \label{fig: data engine pipeline}
    \vspace{-0.5cm}
\end{figure}

\textbf{Overview.}
As shown in Fig~\ref{fig: data engine pipeline} (a), our annotation pipeline consists of 5 stages from data preparation to high-quality QA generation. In stage 1, we employ 3 human experts to manually collect snapshots of 3D point cloud scans from 4 to 6 different directions, with over \textbf{38 hours} of human expertise to maintain the high quality.  In stage 2, we combine the collected snapshots and video frames to generate the structural data that describes different aspects of information in the scene. In stage 3, we leverage GPT-4o to generate over 10k Basic QA covering atomic capabilities with former structural data. In stage 4, human experts again manually filter the low-quality QA pairs, costing over \textbf{112 hours} of 3 experts and resulting in over 8k+ QA pairs. Finally, in stage 5, we design a template-based strategy to integrate the spatial capabilities in QA types, yielding the final QA pairs. The ablation study on the effectiveness of the annotation pipeline can be found at the Sec.~\ref{sec:pipeline effectiveness}.

\textbf{Structural Data Generation.}
As shown in Fig.~\ref{fig: data engine pipeline} (b), this pipeline follows a progressive design to generate structural data with 6 steps:
(1) Initially, 10 keyframes are selected from the video of each scenario by combining the CLIP vision encoder~\cite{clip} and the k-means algorithm.
(2) Based on 2D keyframes, we leverage GPT-4o to generate a 2D caption for each scene, which covers information of appearance, size, and spatial relationships.
(3) We reuse GPT-4o as an inspector to refine the 2D captions by removing incorrect and redundant information.
(4) The manually collected 3D snapshots will be combined with keyframes for 3D caption generation. These high-quality snapshots contain rich global information about the whole scenes, which can provide considerable scene-level spatial information.
(5) The inspector again checks and refines the 3D caption. 
(6) Finally, the rule-based extractor will be applied to obtain structural data for the following QA generation.

\textbf{QA Generation.}
For QA generation, we adopt 3 approaches: template-based, MLLM-based, and human-based generation. For EQ, we use template-based method, and SP is manually designed by 2 human expert. For rest QA types, we leverage GPT-4o to generate. Notably, as we mentioned earlier, the questions in SpaCE-10 are composed of multiple atomic capabilities. However, such highly integrated questions are difficult for current MLLMs to generate directly. Therefore, for SA, OO, OS, EP, FR five QA types, we propose to first generate a basic version of QA, namely basic QA, and then enhance its embedded capabilities. The details of QA generation is in Sec.~\ref{sec:data_generation}.

\textbf{Cross-Capability Integration Strategy.} 
We apply three strategies to integrate cross-capabilities:
(1) For SA, OO, and OS, we integrate an additional C7 (Multi-view Fusion). In the original setting, the four options usually refer to the same object (or object pair). We regroup multiple same-type, same-scene QAs into a single QA so that each option points to different objects. This forces MLLMs to search across the entire scene to find all mentioned entities, enabling a more holistic spatial perception.
(2) For EP, we add C7 and C9 (Reverse Reasoning), expanding the question to involve multiple objects. We then reverse the question type from `which object exists' to `which object does not exist', integrating reverse reasoning capabilities.
(3) For FR, we add C7, C9, and C10 (Situated Observation). The basic FR question asks `which option correctly describes the function of an object near \textit{central object}'. The four options are structured as `Object' which can be used to `Function'. In the integrated version, the question is simply revised to: `Which of the following is correct?'. But Each option involves a different central object, with the structure: `Among the objects adjacent to \textit{central object}, there is one can be used to `Function', but lacks two objects can be used to `Function'. This change prevents potential leakage of prior knowledge in options, such as object names, and places greater emphasis on the model's understanding of functional roles. The examples are in the Sec.~\ref{basic qa and compositional qa}.

\subsection{Quality Verification}
For the quality verification process, we rely on manual filtering by 3 human experts. We set up an user interface for validation and employed two human experts to perform the evaluation. In this process, the evaluation criteria include checking for incorrect options, invalid answers, missing data, or questions involving objects not present in the snapshots. The low-quality data will be directly deleted. This quality control process takes over 112 hours to check and filter low-quality QA pairs. By employing human validation, we ensure that only high-quality and contextually accurate questions are retained for the final benchmark. Related visualizations are attached in the Sec.~\ref{Case Study}. 

\section{Experiments}
\subsection{Setup}
In our experiments, we test nearly 50 close-source and open-source MLLMs on SpaCE-10, including 3D MLLMs LEO and GPT4Scene, GPT-5, InternVL3.5, and so on.
During the evaluation, except for Leo, which is evaluated by their own framework, other MLLMs are evaluated by the LMMs-Eval~\cite{lmms-eval} with 8 frames as input, and open-source MLLMs are tested on Tesla A100 GPUs.  For the response and answer alignment, we follow the prompt of MMBench~\cite{mmbench} and use GPT-4o-2024-11-20 for the judgment. Notably, we removed the random assignment in LMMs-Eval, so the model's performance is likely to be lower than the random baseline (25\%).
\subsection{Overall Results}
\textbf{Human \textit{vs.} MLLMs.}
We first compare human and MLLMs' performance on SpaCE-10. The `Human' score is taken from the average score of 6 human experts. The results in Tab.~\ref{table: main-result} indicate that although human performance does not meet expectations when facing with more complex numerical and reasoning tasks, the total score of 91.2\% is still significantly higher than all existing MLLMs. In comparison, the best open-source MLLM only achieves an average score of 55.0\%, and the best close-source models only achieve a score of 53.1\%. 
These results demonstrate that the compositional spatial intelligence of MLLMs is far below the human level.

\textbf{3D MLLMs \textit{vs.}  2D MLLMs.}
In  Tab.~\ref{table: main-result}, we evaluate LEO and GPT4Scene as representatives of 3D-related MLLMs. Notably, input to the LEO model requires point clouds of the objects relevant to the questions. To ensure a fair comparison, we adjust the input for LEO by randomly sampling 1024 points from the entire scene's point clouds.  
The results show that LEO scored 11.1\% overall in SpaCE-10, which is significantly lower than GPT4Scene (34.5\%). Compared to MLLMs with scale $\leq$ 7B, the performance of LEO-7B is also substantially lower. We argue that one of the limitations of current 3D MLLMs is that they are designed to focus on specific objects and have difficulty processing the entire scene's point clouds as input. Additionally, they likely sacrifice multimodal conversational abilities for understanding scans. These results also indicate that 2D MLLMs have greater potential in visual spatial intelligence comprehension than 3D MLLMs.
\textbf{Open-Source \textit{vs.}  Close-Source.}
In close-souce MLLMs experiments, GPT-5 achieves the best overall performance, ranking 8th with a score of 53.2\%. It excels in perception tasks, such as in SA (Size Assessment) with 69.7\%, OO (Object-Object Spatial Relationship) with 60.7\%, and FR (Functional Reasoning) with 66.8\%, indicating strong accuracy in recognizing size and position. Additionally, GPT-4o achieves the highest score in EQ (Entity Quantification) among all tested models, with a score of 58.3\%. However, GPT-4o struggles with SP, where it scores the lowest among all tasks, suggesting a limitation in scene-level planning ability.  In open-source MLLMs, among models with a scale of over 72B, InternVL3.5-241B-A28B delivers outstanding performance, ranking first overall with a score of 55.0\%. It outperforms other models in nearly all tasks. 
These results suggest that the gap between open-source and close-source models has been significantly narrowed, and some open-source MLLMs even outperform close-source models, especially in compositional spatial intelligence. 
\begin{table}[!t]
\vspace{-0.2cm}
  \setlength{\tabcolsep}{8.3pt}
  \scriptsize
  \centering
  \caption{\textbf{Single-answer performance ranking of nearly 50 MLLMs on SpaCE-10 benchmark.}}
  \vspace{2mm}
  \renewcommand{\arraystretch}{0.95}
  \resizebox{\linewidth}{!}{
\begin{tabular}{@{}lcccccccccc@{}}
    \toprule
    \multirow{3}{*}{\textbf{Models}} & \multirow{3}{*}{\textbf{Rank}} & \multicolumn{5}{c}{\textbf{Perception}} & \multicolumn{3}{c}{\textbf{Reasoning}} & \multirow{3}{*}{\textbf{Overall}} \\
    \cmidrule(lr){3-7} \cmidrule(lr){8-10}
    & & \textbf{EQ} & \textbf{SQ} & \textbf{SA} & \textbf{OO} & \textbf{OS} & \textbf{EP} & \textbf{FR} & \textbf{SP} & \\
    \midrule
    Human &\cellcolor{red!15} 1 & 91.3 & 88.5 & 90.2 & 93.4 & 95.6 & 91.1 & 90.3 & 86.3 & 91.2 \\
    \midrule
    \multicolumn{11}{c}{\tiny{\textbf{3D MLLMs}}} \\
    \midrule
    LEO-7B~\cite{leo} & 47 & 15.8 & 0.0 & 16.7 & 16.5 & 25.2 & 5.5 & 5.7 & 13.3 & 11.1 \\
    GPT4Scene-7B~\cite{gpt4scene} & 36 & 30.9 & 37.7 & 38.0 & 38.9 & 41.6 & 29.5 & 28.0 & 32.5 & 34.5 \\
    \midrule
    \multicolumn{11}{c}{\tiny{\textbf{Close Source 2D MLLMs}}} \\
    \midrule
    GPT-5~\cite{GPT-5} & \cellcolor{yellow!15}3 & 42.0 & 43.0 & 71.0 & 60.7 & 36.5 & 50.3 & 66.8 & 36.0 & 53.4 \\
    GPT-4o-2024-11-20~\cite{gpt4} & 9 & 58.3 & 32.8 & 56.2 & 58.3 & 56.2 & 41.6 & 52.2 & 23.7 & 49.0 \\
    Gemini-2.0-Flash-Thinking~\cite{gemini} & 20 & 34.3 & 25.6 & 53.1 & 42.6 & 53.8 & 42.2 & 46.7 & 31.2 & 42.2 \\
    Claude-3.7-Sonnet~\cite{claude3}& 14 & 46.0 & 44.3 & 49.1 & 46.0 & 49.1 & 44.3 & 49.3 & 25.0 & 46.2 \\
    \midrule
    \multicolumn{11}{c}{\tiny{\textbf{Open Source 2D MLLMs}}} \\
    \midrule
    \rowcolor{gray!15} \multicolumn{11}{l}{$\blacktriangledown$ \emph{Scale $<7$B}} \\
    InternVL2.5-1B~\cite{internvl2.5} & 33 & 33.0 & 54.1 & 18.8 & 43.6 & 29.9 & 26.7 & 41.0 & 23.7 & 35.3 \\
    InternVL3-1B~\cite{internvl3} & 24 & 30.7 & 55.7 & 27.9 & 44.6 & 31.6 & 47.8 & 41.9 & 30.0 & 41.4 \\
    InternVL3.5-1B~\cite{internvl3.5} & 38 & 34.8 & 41.7 & 29.4 & 42.7 & 25.9 & 21.9 & 40.2 & 33.8 & 33.5 \\
    InternVL2.5-2B~\cite{internvl2.5} & 42 & 32.2 & 26.8 & 27.0 & 36.6 & 28.8 & 21.7 & 48.2 & 36.2 & 31.4 \\
    InternVL3-2B~\cite{internvl3} & 17 & 41.5 & 45.9 & 45.4 & 45.7 & 31.9 & 45.7 & 48.7 & 41.3 & 44.2 \\
    InternVL3.5-2B~\cite{internvl3.5} & 35 & 35.6 & 28.4 & 42.2 & 45.7 & 32.3 & 20.1 & 45.8 & 20.0 & 34.6 \\
    Qwen2.5-VL-3B-Instruct~\cite{qwen2.5-vl} & 34 & 31.7 & 23.3 & 47.1 & 51.7 & 31.6 & 25.5 & 37.0 & 21.2 & 34.8 \\
    SpaceOm\textsuperscript{$\diamondsuit$} & 40 & 21.8 & 24.5 & 47.3 & 49.7 & 32.7 & 21.9 & 36.7 & 25.0 & 33.2 \\
    SpaceQwen\textsuperscript{$\diamondsuit$} & 32 & 31.2 & 26.1 & 41.2 & 52.3 & 35.2 & 28.4 & 36.4 & 22.5 & 35.4 \\
    SpaceThinker\textsuperscript{$\diamondsuit$} & 37 & 32.7 & 22.4 & 46.7 & 50.5 & 33.4 & 22.4 & 36.9 & 24.2 & 34.1 \\
    VILA1.5-3B~\cite{vila} & 45 & 25.0 & 9.1 & 31.7 & 34.6 & 31.6 & 35.3 & 12.9 & 33.7 & 26.1 \\
    InternVL2.5-4B~\cite{internvl2.5} & 29 & 34.3 & 23.4 & 50.2 & 50.8 & 16.2 & 21.7 & 56.0 & 33.7 & 35.9 \\
    MiniCPM-v4-4B~\cite{minicpm} & 27 & 38.1 & 32.7 & 41.1 & 49.0 & 36.5 & 29.3 & 50.0 & 30.0 & 39.0 \\
    InternVL3.5-4B~\cite{internvl3.5} & 30 & 38.9 & 12.9 & 48.7 & 50.7 & 27.9 & 33.9 & 37.0 & 35.0 & 35.5 \\
    \rowcolor{gray!15} \multicolumn{11}{l}{$\blacktriangledown$ \emph{Scale $\leq14$B}} \\
    Qwen2.5-VL-7B-Instruct~\cite{qwen2.5-vl} & 39 & 32.7 & 36.9 & 36.9 & 35.3 & 32.3 & 27.6 & 34.2 & 27.5 & 33.3 \\
    LLaVA-v1.5-7B~\cite{llava-v1.5} & 43 & 31.2 & 31.3 & 30.5 & 35.7 & 22.9 & 10.7 & 57.4 & 32.5 & 30.7 \\
    LLaVA-OneVision-7B~\cite{llava-onevision} & 15 & 37.4 & 33.8 & 46.4 & 57.3 & 34.5 & 43.3 & 61.6 & 21.2 & 45.2 \\
    MiMo-VL-RL-8B~\cite{mimovl} & 31 & 23.7 & 35.0 & 46.4 & 41.3 & 34.7 & 32.2 & 32.5 & 36.1 & 35.5 \\
    Cambrian-8B~\cite{cambrian} & 44 & 22.6 & 18.6 & 34.8 & 32.6 & 32.3 & 25.1 & 41.4 & 23.7 & 29.5 \\
    VILA1.5-8B~\cite{vila} & 46 & 25.7 & 8.2 & 27.5 & 32.7 & 17.2 & 12.4 & 26.7 & 23.7 & 20.9 \\
    InternVL2.5-8B~\cite{internvl2.5} & 21 & 33.2 & 36.0 & 50.0 & 55.0 & 33.6 & 27.1 & 59.1 & 32.5 & 41.8 \\
    InternVL3-8B~\cite{internvl3} & 25 & 36.6 & 29.5 & 42.9 & 51.7 & 34.5 & 26.6 & 60.6 & 37.5 & 40.0 \\
    InternVL3.5-8B~\cite{internvl3.5} & 26 & 37.1 & 28.5 & 61.7 & 49.8 & 35.4 & 17.6 & 54.8 & 36.3 & 39.5 \\
    Gemma3-12B~\cite{gemma3} & 22 & 41.8 & 41.2 & 55.1 & 46.5 & 35.6 & 25.0 & 53.2 & 27.5 & 41.5 \\
    InternVL3-14B~\cite{internvl3} & 12 & 39.7 & 28.7 & 54.4 & 58.1 & 38.1 & 51.3 & 56.6 & 35.0 & 47.3 \\
    InternVL3.5-14B~\cite{internvl3.5} & 10 & 41.0 & 47.6 & 65.3 & 52.1 & 34.5 & 45.4 & 54.3 & 30.0 & 48.8 \\
    \rowcolor{gray!15} \multicolumn{11}{l}{$\blacktriangledown$ \emph{14B$<$Scale$<72$B}} \\
    InternVL3.5-20B-A4B~\cite{internvl3.5} & 8 & 37.4 & 43.1 & 64.1 & 58.7 & 41.4 & 54.1 & 57.6 & 28.8 & 51.6 \\
    InternVL2.5-26B~\cite{internvl2.5} & 19 & 34.3 & 29.3 & 62.6 & 55.4 & 33.0 & 29.2 & 61.8 & 33.7 & 43.3 \\
    Gemma3-27B~\cite{gemma3} & 23 & 39.4 & 21.7 & 63.5 & 48.5 & 37.8 & 33.2 & 51.5 & 30.0 & 41.5 \\
    Qwen2.5-VL-32B-Instruct~\cite{qwen2.5-vl} & 41 & 19.9 & 26.5 & 48.9 & 36.8 & 32.3 & 31.1 & 30.1 & 32.5 & 32.6 \\
    InternVL2.5-38B~\cite{internvl2.5} & 16 & 38.1 & 36.1 & 64.4 & 54.3 & 36.8 & 27.4 & 63.0 & 37.5 & 45.1 \\
    InternVL3-38B~\cite{internvl3} & \cellcolor{orange!15}4 & 36.3 & 41.6 & 69.5 & 60.1 & 36.3 & 58.6 & 60.8 & 35.0 & 53.1 \\
    InternVL3.5-38B~\cite{internvl3.5} & 18 & 42.3 & 28.4 & 62.8 & 59.1 & 37.6 & 25.4 & 59.8 & 28.8 & 43.9 \\
    \rowcolor{gray!15} \multicolumn{11}{l}{$\blacktriangledown$ \emph{Scale $\geq72$B}} \\
    GLM-4.5V~\cite{glm45v} & 7 & 38.9 & 41.1 & 65.5 & 61.1 & 36.7 & 61.2 & 49.3 & 31.3 & 51.6 \\
    LLaVA-OneVision-72B~\cite{llava-onevision} & 5 & 44.1 & 38.3 & 67.9 & 64.5 & 40.3 & 46.7 & 67.3 & 36.2 & 52.8 \\
    Qwen2.5-VL-72B-Instruct~\cite{qwen2.5-vl} & 28 & 32.4 & 34.9 & 55.7 & 40.9 & 32.1 & 36.5 & 38.0 & 33.7 & 38.7 \\
    InternVL2.5-78B~\cite{internvl2.5} & 13 & 27.8 & 45.0 & 62.4 & 64.4 & 40.3 & 23.7 & 67.3 & 40.0 & 47.1 \\
    InternVL3-78B~\cite{internvl3} & 6 & 36.8 & 48.2 & 65.3 & 61.6 & 43.8 & 44.4 & 64.3 & 46.3 & 52.5 \\
    Qwen3-VL-235B-A22B-Instruct & 11 & 37.3& 29.6 & 66.4 & 62.9 & 37.8 & 38.6 & 64.0 & 26.8 & 47.9 \\
    InternVL3.5-241B-A28B~\cite{internvl3.5} & \cellcolor{green!15}2 & 35.8 & 39.1 & 68.2 & 63.5 & 46.2 & 64.2 & 58.6 & 40.0 & 55.0 \\
    \bottomrule
\end{tabular}}
\vspace{-0.5cm}
  \label{table: main-result}
{\raggedright \scriptsize\textsuperscript{$\diamondsuit$} Models proposed by RemyxAI (SpaceVLMs series, https://huggingface.co/remyxai/SpaceQwen2.5-VL-3B-Instruct).\par}
\end{table}

\textbf{Single-Answer \textit{vs.}  Multiple-Answer.}
In this experiment, we choose two close MLLM series of similar performance to make a comparison on single-answer and multiple-answer questions. The results in Tab.~\ref{basic vs integrated} show that MLLMs perform significantly worse on multiple-answer tasks compared to single-answer tasks. Smaller models like InternVL2.5-1B and 2B score over 30.0\% on single-answer tasks, while in multiple-answer tasks, their scores often fall below 5\%, which is even worse than random selection. 
As the model size increases to 78B,  MLLMs show greater robustness for different QA types. These results lead to an interesting preliminary conclusion: smaller models may overfit to the single-answer task format, while larger models seem to have learned more fundamental compositional spatial intelligence. However, the Qwen series exhibits an almost opposite trend, with high scores in the smaller models, reaching 0.46. As the parameters increase, the scores decrease, but overall, the performance remains normal.
\begin{table}[!t]
  \scriptsize
\vspace{-0.38cm}
  \centering
  \caption{\textbf{Performance of MLLMs on single-choice and multiple-choice QA pairs. }Results show that MLLMs, especially smaller ones, tend to overfit to single-choice questions.
  }
  \vspace{1mm}
  \renewcommand{\arraystretch}{1.08}
  \resizebox{\linewidth}{!}{
\begin{tabular}{@{}l|ccccc|ccccc|cc|c@{}}
    \toprule
    \multirow{2}{*}{\textbf{Models}} 
    & \multicolumn{5}{c|}{\textbf{Single-choice (3091)}} 
    & \multicolumn{5}{c|}{\textbf{Multiple-choice (970)}} 
    & \multicolumn{2}{c|}{\textbf{Overall\,$\uparrow$}} 
    &\multicolumn{1}{c}{\textbf{Score}\,$\uparrow$}\\
    \cmidrule(lr){2-6} \cmidrule(lr){7-11} \cmidrule(lr){12-13} 
    &  SA & OO & OS & EP & FR 
    &  SA & OO & OS & EP & FR & Single & Multiple
    &  multiple$\big/$single \\
    \midrule
    \multicolumn{14}{c}{\cellcolor{gray!20}\tiny{\textbf{InternVL2.5 Series}}} \\
    InternVL2.5-1B~\cite{internvl2.5} & 18.8 & 43.6 & 29.9 & 26.8 & 41.0 & 4.4 & 3.9 & 4.7 & 0.5 & 11.4 & 32.3 & 3.0 & 0.09 \\
    InternVL2.5-2B~\cite{internvl2.5} & 27.0 & 36.6 & 28.8 & 21.7 & 48.0 & 4.7 & 2.0 & 1.3 &1.5 &  3.5 & 32.4 & 4.1 & 0.13 \\
    InternVL2.5-8B~\cite{internvl2.5} &  50.0 & 55.0 & 33.6 & 27.1 &59.1 & 8.4 & 14.8 & 8.7 & 12.9 & 1.5 & 45.0 & 10.6 & 0.24 \\
    InternVL2.5-38B~\cite{internvl2.5} & 64.4 & 54.3 & 36.8 &  27.4 &63.0 & 47.8& 37.9 & 7.3 & 21.3 & 46.8 & 49.2 & 32.2 & 0.65\\
    InternVL2.5-78B~\cite{internvl2.5} & 62.4 & 64.4 & 40.3 & 23.7 & 67.3 & 38.4 & 33.5 & 12.1 & 12.9& 45.8 &51.6 & 28.5 & 0.55\\
    \multicolumn{14}{c}{\cellcolor{gray!20}\tiny{\textbf{Qwen2.5-VL Series}}} \\
    Qwen2.5-VL-3B-Instruct~\cite{qwen2.5-vl} & 47.1 & 51.7 & 31.6 & 25.5 & 37.0 & 21.7 & 24.1 & 8.0 & 15.3 & 20.7 & 38.6 & 17.9 & 0.46 \\
    Qwen2.5-VL-7B-Instruct~\cite{qwen2.5-vl} & 36.9 & 35.3 & 32.3 & 27.6 & 34.2 & 20.2 & 11.8 & 14.7 & 13.4 & 10.4 & 33.3 & 14.1 & 0.42 \\
    Qwen2.5-VL-32B-Instruct~\cite{qwen2.5-vl} & 48.9 & 36.8 & 32.3 & 31.1 & 30.1 & 25.6 & 6.9 & 8.0 & 10.4 & 12.9 & 35.8 & 12.8 & 0.36 \\
    Qwen2.5-VL-72B-Instruct~\cite{qwen2.5-vl} & 55.7 & 40.9 & 32.1 & 36.5 & 38.0 & 17.2 & 13.3 &13.0 & 15.7 & 15.9 & 40.6 & 15.0 & 0.37 \\
    \multicolumn{14}{c}{\cellcolor{gray!20}\tiny{\textbf{Others}}} \\
    Cambrian-8B~\cite{cambrian} & 34.8 & 32.6 & 32.3 & 25.1 & 41.4 & 5.9 & 12.8 & 5.9 & 0.5 & 20.6 & 33.2 & 9.1 & 0.27 \\
    VILA1.5-8B~\cite{vila} & 27.5 & 32.7 & 17.2 & 12.4 & 26.7 & 0.0 & 0.5 & 1.5 & 2.0 & 26.9 & 23.3 & 6.2 & 0.27 \\
    LLaVA-OneVision-72B~\cite{llava-onevision} &67.9&64.5&40.3&46.7&67.3&38.9&32.0&13.3&34.7&35.8&57.3 &30.9&0.54 \\
    \bottomrule
\end{tabular}}
\vspace{-0.2cm}
  \label{basic vs integrated}
\end{table}

\subsection{Capability Analysis}
\textbf{Atomic Capability \textit{vs.} Compositional Capabilities.}
Firstly, the Tab.~\ref{basic vs integrated} shows that on the 5 types of questions with more compositional abilities, the performance of the four models all drops significantly. 
For the 3 questions with integrating C7 (Multi-view Fusion) capability, the models' accuracy in SA, OO, and OS tasks decreases by 19.1\%, 7.6\%, and 21.0\%, respectively. Despite the drop, the accuracy of SA and OO still remains at a decent level.
Secondly, in the EP task with both C7 and C9 (Reverse Reasoning) abilities integrated, the models' overall accuracy plummets from 67.8\% without integration to 26.8\%, a huge drop of 41.0\%. Similarly, for the FR task, which with the most compositional capacities with an additional C10 (Situated Observation), the four models' average score crashes from 83.0\% to 42.8\%, an even larger drop of 42.9\%.

These results partially reveal the relationship between accuracy and capability. As more abilities are incorporated, model performance declines to varying degrees, with some decreases even exceeding 50.0\%. This indicates that current models have a limited grasp of integrated spatial intelligence, thereby highlighting the necessity of SpaCE-10.

\begin{table}[!t]
\vspace{-0.4cm}
  \scriptsize
  \caption{\textbf{Accuracy comparison on basic and compositional QA pairs.} The results reveal the relationship between the performance and integration of capacities.}
  \vspace{1mm}
  \centering
     \resizebox{\linewidth}{!}{
  \begin{tabular}{@{}l|cccccc|c}
    \toprule
    \multirow{1}{*}{\textbf{Task}} &\textbf{Integrated} & \textbf{Capability} & \textbf{InternVL2.5-1B} & \textbf{ InternVL2.5-8B} & \textbf{Qwen2.5-VL-3B} & \textbf{ Qwen2.5-VL-7B} & \textbf{Average-C(\%)} \\
    \midrule
   
    \multirow{2}{*}{\textbf{SA}} & $\times$& C1,C2,C4,C8& 37.8 & 66.4 & 60.0 & 64.8 & 57.3\\
       &   \cellcolor{gray!20}$\checkmark$  &  \cellcolor{gray!20} +\textbf{C7}&\cellcolor{gray!20} 18.8 ($\downarrow$ 19.0) & \cellcolor{gray!20} 50.0 ($\downarrow$ 16.4) & \cellcolor{gray!20} 47.1 ($\downarrow$ 12.9) & \cellcolor{gray!20} 36.9 ($\downarrow$ 27.9) & \cellcolor{gray!20} 38.2 ($\downarrow$ 19.1)\\
    \midrule
    \multirow{2}{*}{\textbf{OO}}&$\times$ &C1,C2,C3,C8,C9,C10 &52.0 & 66.4 & 56.4 & 41.2 & 54.0 \\
     &  \cellcolor{gray!20} $\checkmark$ & \cellcolor{gray!20}+\textbf{C7}&\cellcolor{gray!20} 43.6 ($\downarrow$ 8.4) & \cellcolor{gray!20} 55.0   ($\downarrow$ 11.4) & \cellcolor{gray!20} 51.7  ($\downarrow$ 4.7) & \cellcolor{gray!20} 35.3   ($\downarrow$ 5.9) &  \cellcolor{gray!20} 46.4   ($\downarrow$ 7.6)\\
    \midrule
    \multirow{2}{*}{\textbf{OS}}& $\times$&C1,C2,C3,C9,C10  &44.2 & 54.8 & 50.8 & 54.3 & 51.0\\
     &  \cellcolor{gray!20} $\checkmark$ & \cellcolor{gray!20} +\textbf{C7}&\cellcolor{gray!20} 29.9 ($\downarrow$ 21.6) & \cellcolor{gray!20} 33.6 ($\downarrow$ 21.2) & \cellcolor{gray!20} 31.6 ($\downarrow$ 19.2) & \cellcolor{gray!20} 32.3 ($\downarrow$ 22.0) &  \cellcolor{gray!20}30.0 ($\downarrow$ 21.0)\\
    \midrule
     \multirow{2}{*}{\textbf{EP}} &$\times$ & C1,C2,C8 &66.6 & 75.8 & 63.4 & 65.3 & 67.8 \\
     & \cellcolor{gray!20}$\checkmark$ & \cellcolor{gray!20}+\textbf{C7},\textbf{C9} &\cellcolor{gray!20} 26.8 ($\downarrow$ 39.8) & \cellcolor{gray!20} 27.1 ($\downarrow$ 48.7) & \cellcolor{gray!20} 25.5 ($\downarrow$ 37.9) & \cellcolor{gray!20} 27.6 ($\downarrow$ 37.7)& \cellcolor{gray!20} 26.8 ($\downarrow$ 41.0) \\
    \midrule
    \multirow{2}{*}{\textbf{FR}} & $\times$& C1,C2,C3,C6,C8&70.9 & 89.7 & 85.6 & 85.7 & 83.0\\
     &  \cellcolor{gray!20}$\checkmark$  & \cellcolor{gray!20} +\textbf{C7},\textbf{C9,C10}&\cellcolor{gray!20} 41.0 ($\downarrow$ 29.9) & \cellcolor{gray!20} 59.1     ($\downarrow$ 30.6) & \cellcolor{gray!20} 37.0  ($\downarrow$ 48.6) & \cellcolor{gray!20} 34.2  ($\downarrow$ 51.5) & \cellcolor{gray!20} 42.8  ($\downarrow$ 42.9)\\
    \bottomrule
  \end{tabular}}
\vspace{-0.5cm}
    \label{basic vs integrated}
\end{table}


\textbf{Spatial Capability Breakdown.} 
To better understand the strengths and weaknesses of the current MLLMs across various atomic capabilities, we constructed a model accuracy and atomic capability score matrix (Fig.~\ref{fig:capability_score_matrix}) to associate QA accuracy with capabilities. Notably, this matrix is an unweighted average, so it diagnoses the model's capability independent of task weights in calculating accuracy. About the calculation details, the QA and capability mapping is shown in Tab.~\ref{tab:task-capability-mapping} and the calculation method is shown in Sec.~\ref{sec:capability}. Moreover, because C1 (Object Recgonition), C7 (Multi-view Fusion), and C8 (Foward Thinking) appear in all eight QA types (100\% coverage), their scores are the model's eight QA-type average accuracies. Our core findings are as follow.

(1) Strength and Weakness:
 C4 (Size Comparison) is consistently the strongest atomic capability across almost all models, whereas  C5 (Counting) is uniformly the weakest. This pattern indicates that continuous magnitude judgments are comparatively well handled, while discrete numerosity under occlusion/clutter and across views remains a persistent perceptual challenge. Notably, while InternVL3.5-241B achieves the highest overall capability (52.3\%) and accuracy (55.0\%) average among all models, its performance on C5 (37.5\%) is substantially lower than that of GPT-4o (45.5\%), whose overall accuracy is 49.0\%. This contrast potentially highlights that closed-source models may possess better generalization in tougher capabilities.

(2) Similar Accuracy $\neq$ Similar Capability: 
Interestingly, comparing GLM-4.5V, LLaVA-OneVision-72B, and InternVL3.5-241B, we observe that their task-weighted overall accuracies are 51.6\%, 52.6\%, and 55.0\%, respectively. Yet GLM's capability average is only 48.1\%, below LLaVA's 52.1\%. Conversely, InternVL3.5-241B's overall accuracy exceeds LLaVA by 2.4\% while its capability average differs by only 0.3\%.
This comparison highlights a crucial fact: high accuracy does not necessarily indicate high capability. Accuracy in Tab.~\ref{table: main-result} emphasizes `task completion rate', whereas the Capability matrix in Fig.~\ref{fig:capability_score_matrix} is another dimension that reveals how MLLMs master diverse spatial capabilities. Relying solely on overall accuracy can therefore mislead about a model's generalizability. The takeaway is that future efforts to boost MLLM spatial intelligence must look beyond accuracy and shore up capabilities, such as C3 (Spatial Relationship), C10 (Situated Observation), and C5 (Counting), to achieve truly robust spatial reasoning. This divergence also validates the necessity of our capability-based analysis.

(3) Scaling improves, but not break bottlenecks: Scaling model parameters under our evaluation framework indeed yields significant overall performance gains. For example, InternVL3.5 improves its capability average from 34.8\% at 2B parameters to 52.3\% at 241B parameters, showing that larger models generally master a broader set of spatial abilities. However, this scaling trend is not uniform across all capabilities. In particular, for C5 (Counting), even the largest InternVL3.5-241B achieves only 37.5\%, far below its strength in other dimensions and still substantially weaker than human performance (89.9\%). This indicates that parameter scaling alone struggles to deliver qualitative improvements in discrete numerical reasoning within spatial contexts. By contrast, models like GPT-4o, GLM-4.5V, and LLaVA-OneVision-72B, achieve superior scores in C5. This suggests that architectural design, training strategy, and data diversity may play a more decisive role than raw scaling for certain atomic spatial capabilities.

\textbf{Spatial Capability Improvement} Taken together, our findings point to two potential pathes for improving spatial intelligence in MLLMs: (1) \textbf{Counting-focused supervision:} curate training sets that stress occlusion, crowding, and multi-view consistency to build reliable discrete counting ability in realistic scenes; (2) \textbf{Capability-aware training beyond scale:} while increasing model size improves overall accuracy and the capability average, gains on the view-conditioned relation chain of C3 (Spatial Relationship) and C10 (Situated Observation) remain limited. Thus, future work could curate training data from a capability perspective, and adopt training strategies (e.g. post-training) based on different capability composition. 

\begin{figure}[!t]
    \centering
    \vspace{-0.5cm}
    \includegraphics[width=0.8\linewidth]{figures/Capability_Score_Matrix_BuPu.png}
    \vspace{-2mm}
    \caption{\textbf{Results of representative MLLMs on 10 atomic capabilities of SpaCE-10.} Each value reflects the model's average accuracy (\%) across all question types involving the respective spatial capability (C1-C10), as defined in the benchmark's task-to-capability mapping.}
    \label{fig:capability_score_matrix}
    \vspace{-0.5cm}
\end{figure}

\section{Conclusion} 
In this paper, we propose SpaCE-10, a comprehensive benchmark for evaluating compositional spatial intelligence in Multimodal Large Language Models (MLLMs). SpaCE-10 covers evaluations of MLLMs from 10 atomic spatial capabilities to 8 compositional capabilities. In SpaCE-10, we collect images and point clouds from  800 scenes and design a hierarchical annotation pipeline to produce over 5\textit{k} high-quality question-answer pairs, covering various evaluation settings of MLLMs.   Through extensive evaluation on nearly 50 MLLMs, we reveal critical limitations of current MLLMs and draw several findings that are beneficial to future work in the community. We believe these studies will provide an invaluable hint for future research toward human-level machine intelligence.





\bibliography{iclr2026_conference}
\bibliographystyle{iclr2026_conference}

\appendix

\section*{Use of Large Language Models (LLMs)}
During the writing of this paper, we utilized LLM solely for language editing to improve clarity and readability. We critically reviewed and revised all AI-generated suggestions to ensure the final text accurately reflects our original intent. All intellectual contributions, including the research design, methodology, analysis, and conclusions, are our exclusive work, and we take full responsibility for the academic integrity of this publication.

\section{Overview of the Appendix}
This appendix provides an introduction of QA definition and examples, atomic capability definition, calculation of correlation between character length and human accuracy, calculation of capability score matrix, experiments on data generation pipeline, case study, annotate interface, ethic statement, and reproducibility statement. It is organized as follows:

In Sec.~\ref{sec:qa_definition}, we detail the definition and example of each QA type. The examples of Basic QA and Compositional QA are also illustrated in this section.
In Sec.~\ref{sec:cap_definition}, we introduce the definition of each atomic spatial capability and our design motivation.
In Sec.~\ref{sec:data_generation}, we describe the detailed process of QA generation in Sec.~\ref{generation of eq and sp}and demonstrate the examples of Basic QA and Compositional QA in Sec.~\ref{basic qa and compositional qa}. The multiple-answer examples are also attached in Sec.~\ref{multi-answer}. 
In Sec.~\ref{sec:capability}, we demonstrate the calculation method of the capability score with a simple example.
In Sec.~\ref{sec:pipeline effectiveness}, we make ablation study on the effectiveness of the data generation pipeline.
In Sec.~\ref{unweighted main results}, we show the unweighted overall performance of Tab.\ref{table: main-result} in the main paper.
In Sec.~\ref{Case Study}, we show the case study of Basic QA quality in Sec.~\ref{Basic QA Quality}, our curated annotation interface in Sec.~\ref{annotation interface}, and more QA cases in Sec.~\ref{visualization QA}. 
In Sec.~\ref{llm usage}, we announce the usage of LLMs.

\section{QA Definition}
\label{sec:qa_definition}
\begin{figure}[!t]
    \centering
    \includegraphics[width=1\linewidth]{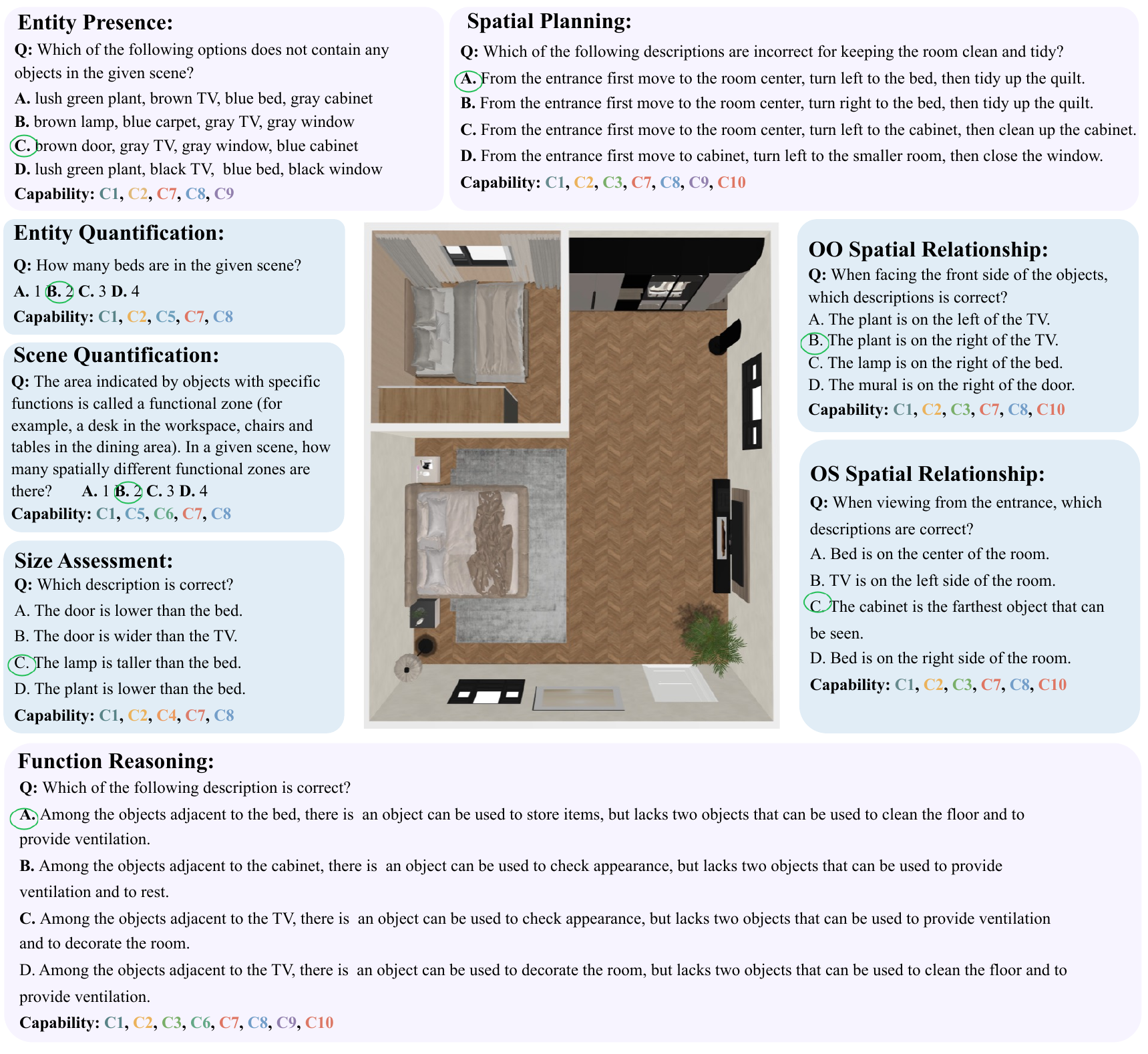}
    \caption{\textbf{Examples of all types of QA.} The blue examples represent the perception QA, and the purple ones denote the reasoning QA. The green circles are the correct answer.}
    \label{fig: all qa}
\end{figure}

In this section, we introduce the definition of each QA type and show the QA examples in Fig.~\ref{fig: all qa}. Notably, each QA type is the integration of multiple capabilities, and we also show the mapping between QA and capabilities in this figure and Tab.~\ref{tab:task-capability-mapping}.
The following are definitions:  \\
\begin{itemize}
    \item Entity Quantification (EQ): Counting the number of objects in a scene.
    \item Scene Quantification (SQ): Counting the number of regions within a scene.
    \item Size Assessment (SA): Comparing the size relationships between different objects.
    \item Object-Object Spatial Relationship (OO): Understanding the relative spatial relationship between two objects.
    \item Object-Scene Spatial Relationship (OS): Understanding the relative spatial relationship between an object and the overall scene.
    \item Functional Reasoning (FR): Reasoning objects that match or do not match certain functions based on relative spatial relationships.
    \item Spatial Planning (SP): Global spatial navigation and path planning.
\end{itemize}

\begin{table}[]
\centering
\caption{\textbf{Mapping of QA types and Spatial Capabilities.} Each type of QA is the integration of multiple capabilities.}
\label{tab:task-capability-mapping}
\renewcommand{\arraystretch}{1.2}
\setlength{\tabcolsep}{8pt} 
\scriptsize
\begin{tabular}{lcccccccccc}
\toprule
Tasks & c1  & c2  & c3      & c4       & c5      & c6     & c7  &  c8 & c9   & c10    \\ \hline
Entity Quantification (EQ)    & $\checkmark$ & $\checkmark$ & -         &  -  & $\checkmark$ & -    &   $\checkmark$& $\checkmark$     & -       & -  \\
Scene Quantification (SQ)   & $\checkmark$ &   -   & -  & -    & $\checkmark$ & $\checkmark$ & $\checkmark$ &        $\checkmark$         &   -   &-     \\
Size Assessment (SA)    & $\checkmark$ & $\checkmark$ &  -           & $\checkmark$ &    -  &    -      & $\checkmark$     & $\checkmark$ &     -         &          -    \\
OO-Spatial Relationship (OO)  & $\checkmark$ & $\checkmark$ & $\checkmark$ &   - &  -& - & $\checkmark$ &  $\checkmark$       &  -  & $\checkmark$ \\
OS-Spatial Relationship (OS)   & $\checkmark$ & $\checkmark$ & $\checkmark$ &   -  &   - &   -& $\checkmark$ & $\checkmark$  & -& $\checkmark$  \\ 
Entity Presence (EP) & $\checkmark$ & $\checkmark$ &- & -&- & -& $\checkmark$ & $\checkmark$ & $\checkmark$&- \\
Functional Reasoning (FR)  & $\checkmark$ & $\checkmark$ & $\checkmark$ &   - &  - & $\checkmark$ & $\checkmark$ & $\checkmark$ & $\checkmark$ & $\checkmark$ \\

Spatial Planning (SP)  & $\checkmark$ & $\checkmark$ & $\checkmark$ &  -   &  -  &  -  & $\checkmark$ &  $\checkmark$  & $\checkmark$ & $\checkmark$\\ \bottomrule
\end{tabular}
\vspace{0.3cm}

\end{table}
%

\section{Capability Definition}
\label{sec:cap_definition}
In this section, we illustrate the definition of each capability:\\
C1-Object Recognition: Identify what the object is.\\
C2-Spatial Location: Localizing an object in space.\\
C3-Spatial Relationship: Understanding relative spatial position relationship.\\
C4-Size Comparison: Comparing the size relationship of objects.\\
C5-Counting: Count the number of objects and scenes.\\
C6-Function Knowledge: Understanding the function of objects.\\
C7-Multi-view Fusion: Understanding spatial information from multiple views\\
C8-Forward Thinking: Understand the forward instructions and complete tasks in the given space.\\
C9-Reverse Reasoning: Understand the reverse instructions and complete tasks in the given space.  \\
C10-Situated Observation: Imagine standing in a designated position in space and observing and understanding the scene.\\

In designing our capabilities, we draw on established research in spatial intelligence. Initially, a beginning reference for our design is Sparkle~\cite{sparkle}, which frames 2D spatial intelligence around three abilities—Direction, Distance, and Location. In SpaCE-10, the direction and distance are split into C3 (Spatial Relationship). Since we believe that the concept of `what and where' is one of the most fundamental abilities for real spatial intelligence, location is separated and mapped to the C2 (Spatial Location). Next, early studies~\cite{whatsup,elephants} mainly emphasize C4 (Size Comparison) and simple positional cues in clean images. Then, as evaluation expands from single-scene 3D point-cloud QA~\cite{scanqa,sqa3d,3dqa,fe-3dgqa,clever3d,3d-llm,3dgrand,mmscan,beacon,reason3d} to multi-view 2D imagery~\cite{3dsrbench,spatialllm,msqa,vsi,mmsi,omnispatial,spatialrgpt}, the focus shifts from local relations (C3) and counting (C5) within one scene to cross-view consistency captured by C7 (Multi-view Fusion). In parallel, SQA~\cite{sqa3d} introduces viewpoint conditioning, aligning with C10 (Situated Observation). Nowadays, with the rise of Embodied Intelligence and Reasoning MLLMs, based on these previous excellent works, we further propose to incorporate C6 (Function Knowledge), C8 (Forward Thinking), and C9 (Reverse Reasoning) into this atomic capability pool for manipulating basic knowledge and reasoning ability examination.

\section{Data Generation Details}
\label{sec:data_generation}
\subsection{Generation for EQ and SP} 
\label{generation of eq and sp}
For EQ, we first extract the number of each object from the scan datasets' semantic labels, and then leverage a predefined template to generate QA. For the SP, we employed human experts to manually design 80 QA pairs. Each question presents a complete task flow, including the navigation path, goal, goal characteristics, and actions to be performed, with potential errors in any step. Based on the prompt, the model must select either a fully correct task flow or one containing incorrect steps. 

\subsection{Basic QA and Compositional QA}
\label{basic qa and compositional qa}
In the main paper, we mention that there are 5 QA types (OO, OS, SA, FR, SP) that will be applied with cross-capability integration strategy. Thus, in this section, we demonstrate examples of these two QAs. 

(1) For OO (Object-Object Spatial Relationship), OS (Object-Scene Spatial Relationship), SA (Size Assessment):

\textbf{Basic-SA-Question:} When facing the front side of the objects, which description is correct?

\textbf{Basic-SA-Options:}

A. This red table is taller than the brown cabinet, but narrower than the brown cabinet.

B. This red table is shorter and narrower than the brown cabinet.

C. This red table is shorter, but wider than the brown cabinet.

D. This red table is taller than the brown cabinet and wider than brown cabinet.

\textbf{Basic-SA-Answer:} B

In the Basic QA format, each question involves only two objects, and the scenario is restricted to a single viewpoint, i.e., the front of the objects. For each scene, we generate multiple Basic QA questions and then aggregate these options to form Compositional QA questions.

\textbf{Compositional-SA-Question:} When facing the front side of the objects, which description is correct?

\textbf{Compositional-SA-Options:}

A. This red table is taller than the brown cabinet, but narrower than brown cabinet.

B. The blue sofa is wider than the white door.

C. The green plant is taller and wider than the wooden bookshelf.

D. This red table is taller than the brown cabinet and wider than brown cabinet.

\textbf{Compositional-SA-Answer:} B

In Compositional QA, we combine multiple Basic QA questions, so each question may now refer to different objects located at various positions across the scene. This forces the model to integrate information from multiple perspectives, requiring C7: Multi-view Fusion.

(2) For EP (Entity Presence), we add additional C7 and C9:

\textbf{Basic-EP-Question 1:} Does a red chair exist in the given scene?

\textbf{Basic-EP-Options 1:}

A. Yes B. No

\textbf{Basic-EP-Answer 1:} A

\textbf{Basic-EP-Question 2:} Does a gray sofa exist in the given scene?

\textbf{Basic-EP-Options 2:}

A. Yes B. No

\textbf{Basic-EP-Answer 2:} B

We generate multiple Basic-EP questions and then aggregate them into Compositional-EP questions.

\textbf{Compositional-EP-Question:} Which of the following options does not contain any objects in the given scene?

\textbf{Compositional-EP-Options:}

A. Red chair, blue sofa, green plant, red table

B. Gray sofa, white lamp, orange carpet, pink cushion

C. Black coffee table, grey armchair, purple curtain, brown bookshelf

D. Beige ottoman, teal vase, silver lamp, golden picture frame

\textbf{Compositional-EP-Answer:} B

In Compositional-EP, each option contains multiple objects. Similar to Basic QA, where objects in the scene may appear from different angles or positions, C7: Multi-view Fusion is required. Additionally, the task shifts from a direct "does this object exist?" question to a more complex "which object is missing from the scene?" requiring C9: Reversed Reasoning.

(3) For FR (Functional Reasoning), we add additional C7, C9, and C10:

\textbf{Basic-FR-Question:} Which option correctly describes the function of an object near the bed?

\textbf{Basic-FR-Options:}

A. Cabinet, store items

B. Mop, clean the floor

C. Window, provide ventilation

D. Mop, provide light

\textbf{Basic-FR-Answer:} A

In Basic-FR, the correct answer must satisfy both functional and spatial positioning requirements. We generate multiple Basic-FR questions, which are then combined into Compositional-FR questions.

\textbf{Compositional-FR-Question:} Which of the following descriptions is correct?

\textbf{Compositional-FR-Options:}

A. Among the objects adjacent to the bed, there is an object that can be used to store items, but lacks two objects that can be used to clean the floor and provide ventilation.

B. Among the objects adjacent to the cabinet, there is an object that can be used to check our appearance, but lacks two objects that can be used to provide ventilation and rest.

C. Among the objects adjacent to the TV, there is an object that can be used to check our appearance, but lacks two objects that can be used to provide ventilation and decorate the room.

D. Among the objects adjacent to the TV, there is an object that can be used to decorate the room, but lacks two objects that can be used to clean the floor and provide ventilation.

\textbf{Compositional-FR-Answer:} A

In Compositional-FR, multiple central objects are combined, broadening the question to encompass the entire scene. The model must analyze the scene from various perspectives, necessitating C7: Multi-view Fusion to integrate spatial relationships and C10: Situated Observation to understand the contextual functionality of objects. Furthermore, the task involves evaluating the relevance of functional descriptions, which requires C9: Reverse Reasoning to assess the appropriateness of the functions in relation to the scene.

\subsection{Multi-Answer Example}
\label{multi-answer}
Each multiple-choice question is set as a five-option, two-correct answer question, where the evaluated model must select two correct answers to be considered correct. The accuracy for each type of QA is calculated in the same way as single-choice questions. Compared to single-choice questions, the question and option format for multiple-choice questions remains exactly the same. Below is an example of the EP:

\textbf{Single-Choice EP:}

Question: Which of the following options does not contain any objects in the given scene?

Options:

A. Red chair, blue sofa, green plant, red table

B. Gray sofa, white lamp, orange carpet, pink cushion

C. Black coffee table, grey armchair, purple curtain, brown bookshelf

D. Beige ottoman, teal vase, silver lamp, golden picture frame

Answer: B

Double-Choice EP:

Question: Which of the following options does not contain any objects in the given scene?

Options:

A. Red chair, blue sofa, green plant, red table

B. Gray sofa, white lamp, orange carpet, pink cushion

C. Black coffee table, grey armchair, purple curtain, brown bookshelf

D. Beige ottoman, teal vase, silver lamp, golden picture frame

E. Gray sofa, grey armchair, purple curtain, brown bookshelf

Answer: B, E

\section{Capability Score Calculaion}
\label{sec:capability}
\paragraph{Capability-score computation (for Fig.~\ref{fig:capability_score_matrix}).}
The values in Fig.~\ref{fig:capability_score_matrix} are computed from the per-question scores in Tab.~\ref{tab:task-capability-mapping} via the task$\to$capability mapping (C1-C10). 

Let $\mathcal{Q}_i$ denote the set of questions linked to capability $C_i$ and $n_i = |\mathcal{Q}_i|$.
For a question $q \in \mathcal{Q}_i$, let $\mathrm{Score}(q)$ be the score assigned by the task's evaluation rule (e.g., 0/100 for single-answer exact match, or the task-specific percentage for multiple-answer).
The capability score is the mean over its linked questions:
\begin{equation}
\mathrm{Score}(C_i) \;=\; \frac{1}{n_i}\sum_{q \in \mathcal{Q}_i} \mathrm{Score}(q).
\label{eq:cap-score}
\end{equation}
If no question maps to a capability ($n_i=0$), the entry is marked as \texttt{N/A} and excluded from any further averaging. 
When a question maps to multiple capabilities, its score contributes to each linked capability (no reweighting).

\paragraph{Toy example.}
Consider two capabilities $C_1$ and $C_2$, and two questions $Q_1,Q_2$ with mappings
$Q_1 \mapsto C_1$ and $Q_2 \mapsto \{C_1,C_2\}$.
Suppose $\mathrm{Score}(Q_1)=80$ and $\mathrm{Score}(Q_2)=90$.
Then $\mathcal{Q}_1=\{Q_1,Q_2\}$ ($n_1=2$) and $\mathcal{Q}_2=\{Q_2\}$ ($n_2=1$), yielding
\[
\mathrm{Score}(C_1)=\tfrac{80+90}{2}=85, \qquad
\mathrm{Score}(C_2)=90.
\]
This procedure produces the capability values shown in Fig.~\ref{fig:capability_score_matrix}.

\section{Pipeline Component Ablation }
\label{sec:pipeline effectiveness}
\begin{table}[]
\centering
\scriptsize
\label{table: pipeline component}
\caption{\textbf{Ablation study of structural data generation pipeline.} We randomly sample 30 scenes for each type of QA and generate one QA for each scene. The results demonstrate the effectiveness of each component.}
\begin{tabular}{lcccccc}
\toprule
Component          & SA   & OO   & OS   & EP   & FP   & Average \\ \midrule
2D Captioner       & 50.0 & 50.0 & 46.7 & 96.7 & 66.7 & 62.0    \\
+ 3D Captioner     & 80.0 & 63.3 & 63.3 & 96.7 & 80.0 & 76.7 (+14.7)   \\
+ Inspector        & 80.0 & 66.7 & 70.0 & 96.7 & 83.3 & 79.3 (+17.3)   \\
+ Structural Data & 86.7 & 76.7 & 76.7 & 96.7 & 83.3 & 84.0 (+22.0)   \\ \bottomrule
\end{tabular}
\end{table}

To systematically evaluate the contribution of each key component in the structured data generation pipeline to the accuracy of basic QA generation, we designed a stepwise ablation study. Specifically, we selected five QA task types (SA, OO, OS, EP, FP), and for each type, we randomly sampled 30 scenes, generating one question per scene. The accuracy of each generated question was manually verified by human experts.

The experimental results in Tab.~\ref{table: pipeline component} clearly demonstrate the cumulative gains of each module. Using the 2D Captioner alone as the baseline, the generated QA already showed relatively stable accuracy across most categories (an average of 62.0\%), with particularly high accuracy in the EP task at 96.7\%. This reflects the relatively low difficulty of generating questions for this category and that 2D visual information sufficiently supports it. With the addition of the 3D Captioner, the overall accuracy improved significantly (an average increase of 14.7\%), indicating that 3D information effectively supplements the limitations of 2D vision and enhances the model's understanding of spatial and object attributes. Further incorporating the Inspector component led to another increase in accuracy, reaching 79.3\% (a 17.3\% improvement over the baseline), showing that this module plays an important role in validating and refining the details of question generation. Finally, after adding structured data, the overall average accuracy reached 84.0\%, a 22\% improvement compared to the initial baseline, fully demonstrating the critical value of structured information in improving the quality of basic QA generation.
\section{Unweighted Main Results}
\label{unweighted main results}
We demonstrate the unweighted single-answer performance ranking on SpaCE-10 in Tab.~\ref{table: macro-supp}. 
\begin{table}[!t]
  \setlength{\tabcolsep}{8.3pt}
\scriptsize
  \caption{\textbf{Unweighted overall performance ranking on SpaCE-10.}}
  \renewcommand{\arraystretch}{1.0}
  \centering
  \resizebox{1\linewidth}{!}{
\begin{tabular}{@{}lcccccccccc@{}}
    \toprule
    \multirow{3}{*}{\textbf{Models}} & \multirow{3}{*}{\textbf{Rank}} & \multicolumn{5}{c}{\textbf{Perception}} & \multicolumn{3}{c}{\textbf{Reasoning}} & \multirow{3}{*}{\textbf{Overall}} \\
    \cmidrule(lr){3-7} \cmidrule(lr){8-10}
    & & \textbf{EQ} & \textbf{SQ} & \textbf{SA} & \textbf{OO} & \textbf{OS} & \textbf{EP} & \textbf{FR} & \textbf{SP} & \\
    \midrule
    Human & 1 & 91.3 & 88.5 & 90.2 & 93.4 & 95.6 & 91.1 & 90.3 & 86.3 & 90.8 \\
    \midrule
    \multicolumn{11}{c}{\tiny{\textbf{3D MLLMs}}} \\
    \midrule
    LEO-7B~\cite{leo} & 46 & 15.8 & 0.0 & 16.7 & 16.5 & 25.2 & 5.5 & 5.7 & 13.3 & 12.3 \\
    GPT4Scene-7B~\cite{gpt4scene} & 31 & 30.9 & 37.7 & 38.0 & 38.9 & 41.6 & 29.5 & 28.0 & 32.5 & 34.6 \\
    \midrule
    \multicolumn{11}{c}{\tiny{\textbf{Close Source 2D MLLMs}}} \\
    \midrule
    GPT-5~\cite{GPT-5} & 5 & 42.0 & 43.0 & 69.7 & 60.7 & 36.5 & 50.3 & 66.8 & 30.0 & 49.9 \\
    GPT-4o~\cite{gpt4} & 9 & 58.3 & 32.8 & 56.2 & 58.3 & 56.2 & 41.6 & 52.2 & 23.7 & 47.4 \\
    Gemini-2.0~\cite{gemini} & 19 & 34.3 & 25.6 & 53.1 & 42.6 & 53.8 & 42.2 & 46.7 & 31.2 & 41.2 \\
    Claude-3.7-Sonnet~\cite{claude3} & 14 & 46.0 & 44.3 & 49.1 & 46.0 & 49.1 & 44.3 & 49.3 & 25.0 & 44.1 \\
    \midrule
    \multicolumn{11}{c}{\tiny{\textbf{Open Source 2D MLLMs}}} \\
    \midrule
    \rowcolor{gray!15} \multicolumn{11}{l}{$\blacktriangledown$ \emph{Scale $<7$B}} \\
    InternVL2.5-1B~\cite{internvl2.5} & 33 & 33.0 & 54.1 & 18.8 & 43.6 & 29.9 & 26.7 & 41.0 & 23.7 & 33.9 \\
    InternVL3-1B~\cite{internvl3} & 25 & 30.7 & 55.7 & 27.9 & 44.6 & 31.6 & 47.8 & 41.9 & 30.0 & 38.8 \\
    InternVL3.5-1B~\cite{internvl3.5} & 34 & 34.8 & 41.7 & 29.4 & 42.7 & 25.9 & 21.9 & 40.2 & 33.8 & 33.8 \\
    InternVL2.5-2B~\cite{internvl2.5} & 41 & 32.2 & 26.8 & 27.0 & 36.6 & 28.8 & 21.7 & 48.2 & 36.2 & 32.2 \\
    InternVL3-2B~\cite{internvl3} & 15 & 41.5 & 45.9 & 45.4 & 45.7 & 31.9 & 45.7 & 48.7 & 41.3 & 43.3 \\
    InternVL3.5-2B~\cite{internvl3.5} & 35 & 35.6 & 28.4 & 42.2 & 45.7 & 32.3 & 20.1 & 45.8 & 20.0 & 33.8 \\
    Qwen2.5-VL-3B-Instruct & 37 & 31.7 & 23.3 & 47.1 & 51.7 & 31.6 & 25.5 & 37.0 & 21.2 & 33.6 \\
    SpaceOm\textsuperscript{$\diamondsuit$} & 39 & 21.8 & 24.5 & 47.3 & 49.7 & 32.7 & 21.9 & 36.7 & 25.0 & 32.5 \\
    SpaceQwen\textsuperscript{$\diamondsuit$} & 32 & 31.2 & 26.1 & 41.2 & 52.3 & 35.2 & 28.4 & 36.4 & 22.5 & 34.2 \\
    SpaceThinker\textsuperscript{$\diamondsuit$} & 36 & 32.7 & 22.4 & 46.7 & 50.5 & 33.4 & 22.4 & 36.9 & 24.2 & 33.6 \\
    VILA1.5-3B~\cite{vila} & 44 & 25.0 & 9.1 & 31.7 & 34.6 & 31.6 & 35.3 & 12.9 & 33.7 & 26.7 \\
    InternVL2.5-4B~\cite{internvl2.5} & 28 & 34.3 & 23.4 & 50.2 & 50.8 & 16.2 & 21.7 & 56.0 & 33.7 & 35.8 \\
    MiniCPM-v4-4B~\cite{minicpm} & 26 & 38.1 & 32.7 & 41.1 & 49.0 & 36.5 & 29.3 & 50.0 & 30.0 & 38.3 \\
    InternVL3.5-4B~\cite{internvl3.5} & 29 & 38.9 & 12.9 & 48.7 & 50.7 & 27.9 & 33.9 & 37.0 & 35.0 & 35.6 \\
    \rowcolor{gray!15} \multicolumn{11}{l}{$\blacktriangledown$ \emph{Scale $\leq14$B}} \\
    Qwen2.5-VL-7B-Instruct~\cite{qwen2.5-vl} & 38 & 32.7 & 36.9 & 36.9 & 35.3 & 32.3 & 27.6 & 34.2 & 27.5 & 32.9 \\
    LLaVA-v1.5-7B~\cite{llava-v1.5} & 42 & 31.2 & 31.3 & 30.5 & 35.7 & 22.9 & 10.7 & 57.4 & 32.5 & 31.5 \\
    LLaVA-OneVision-7B~\cite{llava-onevision} & 18 & 37.4 & 33.8 & 46.4 & 57.3 & 34.5 & 43.3 & 61.6 & 21.2 & 41.9 \\
    MiMo-VL-RL-8B~\cite{mimovl} & 30 & 23.7 & 35.0 & 46.4 & 41.3 & 34.7 & 32.2 & 32.5 & 36.1 & 35.2 \\
    Cambrian-8B~\cite{cambrian} & 43 & 22.6 & 18.6 & 34.8 & 32.6 & 32.3 & 25.1 & 41.4 & 23.7 & 28.9 \\
    VILA1.5-8B~\cite{vila} & 45 & 25.7 & 8.2 & 27.5 & 32.7 & 17.2 & 12.4 & 26.7 & 23.7 & 21.8 \\
    InternVL2.5-8B~\cite{internvl2.5} & 20 & 33.2 & 36.0 & 50.0 & 55.0 & 33.6 & 27.1 & 59.1 & 32.5 & 40.8 \\
    InternVL3-8B~\cite{internvl3} & 24 & 36.6 & 29.5 & 42.9 & 51.7 & 34.5 & 26.6 & 60.6 & 37.5 & 40.0 \\
    InternVL3.5-8B~\cite{internvl3.5} & 23 & 37.1 & 28.5 & 61.7 & 49.8 & 35.4 & 17.6 & 54.8 & 36.3 & 40.1 \\
    Gemma3-12B~\cite{gemma3} & 21 & 41.8 & 41.2 & 55.1 & 46.5 & 35.6 & 25.0 & 53.2 & 27.5 & 40.7 \\
    InternVL3-14B~\cite{internvl3} & 12 & 39.7 & 28.7 & 54.4 & 58.1 & 38.1 & 51.3 & 56.6 & 35.0 & 45.2 \\
    InternVL3.5-14B~\cite{internvl3.5} & 11 & 41.0 & 47.6 & 65.3 & 52.1 & 34.5 & 45.4 & 54.3 & 30.0 & 46.3 \\
    \rowcolor{gray!15} \multicolumn{11}{l}{$\blacktriangledown$ \emph{14B$<$Scale$<72$B}} \\
    InternVL3.5-20B-A4B~\cite{internvl3.5} & 7 & 37.4 & 43.1 & 64.1 & 58.7 & 41.4 & 54.1 & 57.6 & 28.8 & 48.2 \\
    InternVL2.5-26B~\cite{internvl2.5} & 17 & 34.3 & 29.3 & 62.6 & 55.4 & 33.0 & 29.2 & 61.8 & 33.7 & 42.4 \\
    Gemma3-27B~\cite{gemma3} & 22 & 39.4 & 21.7 & 63.5 & 48.5 & 37.8 & 33.2 & 51.5 & 30.0 & 40.7 \\
    Qwen2.5-VL-32B-Instruct~\cite{qwen2.5-vl} & 40 & 19.9 & 26.5 & 48.9 & 36.8 & 32.3 & 31.1 & 30.1 & 32.5 & 32.3 \\
    InternVL2.5-38B~\cite{internvl2.5} & 13 & 38.1 & 36.1 & 64.4 & 54.3 & 36.8 & 27.4 & 63.0 & 37.5 & 44.7 \\
    InternVL3-38B~\cite{internvl3} & 6 & 36.3 & 41.6 & 69.5 & 60.1 & 36.3 & 58.6 & 60.8 & 35.0 & 49.8 \\
    InternVL3.5-38B~\cite{internvl3.5} & 16 & 42.3 & 28.4 & 62.8 & 59.1 & 37.6 & 25.4 & 59.8 & 28.8 & 43.0 \\
    \rowcolor{gray!15} \multicolumn{11}{l}{$\blacktriangledown$ \emph{Scale $\geq72$B}} \\
    GLM-4.5V~\cite{glm45v} & 8 & 38.9 & 41.1 & 65.5 & 61.1 & 36.7 & 61.2 & 49.3 & 31.3 & 48.1 \\
    LLaVA-OneVision-72B~\cite{llava-onevision} & 4 & 44.1 & 38.3 & 67.9 & 64.5 & 40.3 & 46.7 & 67.3 & 36.2 & 50.7 \\
    Qwen2.5-VL-72B-Instruct~\cite{qwen2.5-vl} & 27 & 32.4 & 34.9 & 55.7 & 40.9 & 32.1 & 36.5 & 38.0 & 33.7 & 38.0 \\
    InternVL2.5-78B~\cite{internvl2.5} & 10 & 27.8 & 45.0 & 62.4 & 64.4 & 40.3 & 23.7 & 67.3 & 40.0 & 46.4 \\
    InternVL3-78B~\cite{internvl3} & 3 & 36.8 & 48.2 & 65.3 & 61.6 & 43.8 & 44.4 & 64.3 & 46.3 & 51.3 \\
    InternVL3.5-241B-A28B~\cite{internvl3.5} & 2 & 35.8 & 39.1 & 68.2 & 63.5 & 46.2 & 64.2 & 58.6 & 40.0 & 52.0 \\
    \bottomrule
\end{tabular}}
  \label{table: macro-supp}
  
{\raggedright \scriptsize\textsuperscript{$\diamondsuit$} Models proposed by RemyxAI (SpaceVLMs series, https://huggingface.co/remyxai/SpaceQwen2.5-VL-3B-Instruct).\par}
\end{table}

\section{Case Study}
\label{Case Study}
\subsection{Basic QA Quality}
\label{Basic QA Quality}
In this paper, we have developed a sophisticated pipeline for the automated generation of high-quality Basic QA (Basic Question Answering). In the previous section, we systematically verified the effectiveness of each component within the pipeline. This section presents two typical high-quality QA cases and delves into them with specific context-based analysis and discussion.

As shown in Fig.~\ref{fig: Basic SA case}, the first case centers on the size estimation task (Basic SA). As depicted in the figure, through meticulously crafted prompts and high-quality snapshots, we steered GPT to accurately generate questions. When comparing the volume relationship between a white wooden baby crib and a wardrobe, GPT delivered an impressive performance. It not only correctly identified the height and width dimensions of each object but also precisely determined their relative differences across multiple perspectives. For instance, its response accurately pointed out that the wardrobe is taller than the crib but has a similar width, demonstrating a good grasp of the 3D geometric properties of objects.

The second case focuses on the spatial relationship understanding task (Basic OO), as shown in the Fig.~\ref{fig: Basic OO case}, in the context of judging the spatial relationship between a yellow chair and an entrance. In this example, GPT not only accurately distinguished the relative direction of "left" but also correctly identified the spatial distance difference between "near" and "far". Although the question only involved one yellow chair, the distractors in the options were somewhat deceptive, which in turn enhanced the question's ability to assess the model's spatial understanding. This case also indirectly confirms the feasibility and rationality of using GPT for generating spatial intelligence QA questions.
Through these two cases, we observed that powerful MLLMs, when only provided with high-quality 2D visual inputs, are already capable of understanding certain 3D spatial information, such as object size, volume relationships, and spatial orientation. This ability suggests a promising path for future exploration: unlike current 3D models that sacrifice conversational abilities to fit point cloud data, 2D MLLMs can still demonstrate strong spatial understanding potential even without explicitly incorporating 3D structural modeling.
\begin{figure}[!t]
    \centering
    \includegraphics[width=1\linewidth]{figures/space-10_1.pdf}
    \caption{\textbf{Showcase of Basic SA QA.} `H' and `W' mean the height and width of objects, respectively. In this case, GPT-4o shows the precise understanding of both size and geometry.}
    \label{fig: Basic SA case}
\end{figure}

\begin{figure}
    \centering
    \includegraphics[width=1\linewidth]{figures/space-10_2.pdf}
    \caption{\textbf{Showcase of Basic OO.} This case reflects the capability in spatial relationships and QA design. Notably, all spatial relationship is based on the observation at the entrance.}
    \label{fig: Basic OO case}
\end{figure}

\begin{figure}
    \centering
    \includegraphics[width=1\linewidth]{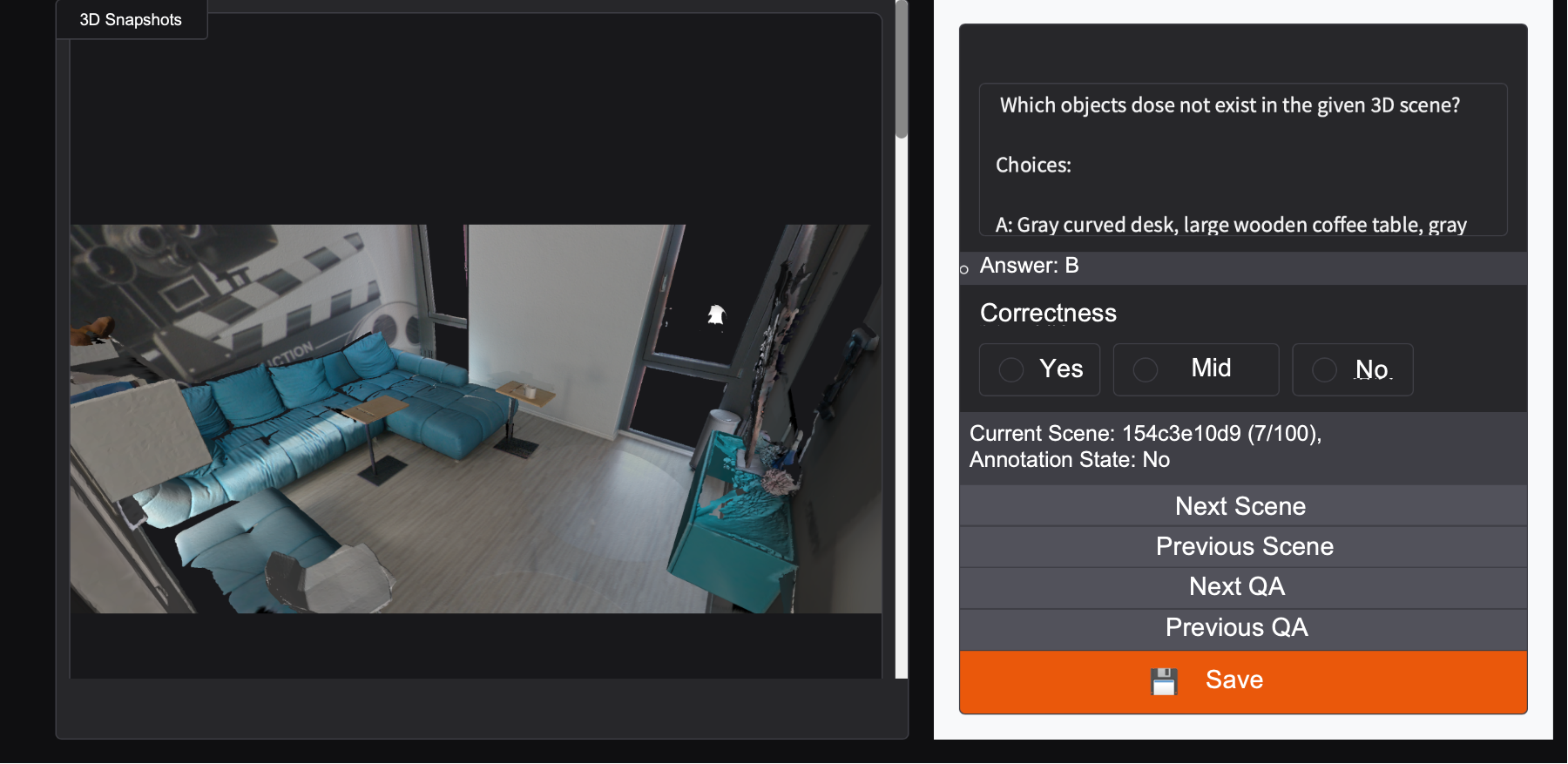}
    \caption{\textbf{Interface of annotation tools.}}
    \label{fig: annotation}
\end{figure}
\subsection{Annotation Interface}
\label{annotation interface}
To facilitate efficient annotation, we developed a custom annotation tool, with the interface shown in Fig.~\ref{fig: annotation}. During the annotation process, human experts are restricted to viewing only 3D snapshots to judge the correctness of each QA pair; they are not allowed to view the 2D images. This design offers two main advantages: (1) It ensures a high annotation speed. By examining only a small number of 3D snapshots, annotators can quickly grasp the overall layout of the scene while significantly reducing their visual workload. (2) Since 2D images contain overly fine-grained details, many of which may not be present in the 3D scene, using only 3D information to filter out incorrect questions helps ensure that the resulting QA pairs are suitable for evaluation across both 2D and 3D models.
Additionally, during evaluation, we tag erroneous questions to ensure none are overlooked. This end-to-end process not only prioritizes annotation efficiency but also reflects our rigorous commitment to data quality control.

\subsection{visualization of QA}
\begin{figure}
    \centering
    \includegraphics[width=0.6\linewidth]{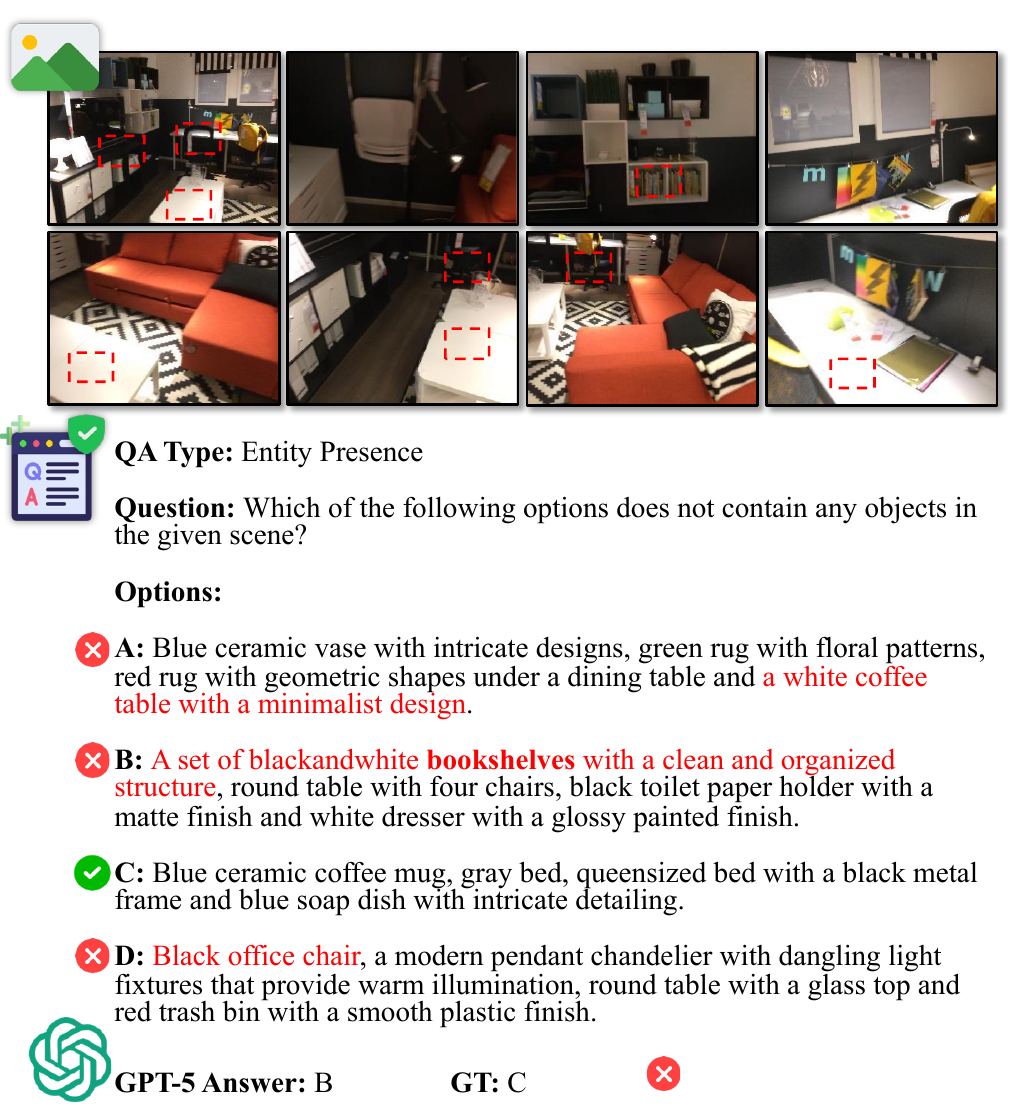}
    \caption{Visualization of Entity Presence}
    \label{ep1}
\end{figure}

\begin{figure}
    \centering
    \includegraphics[width=0.6\linewidth]{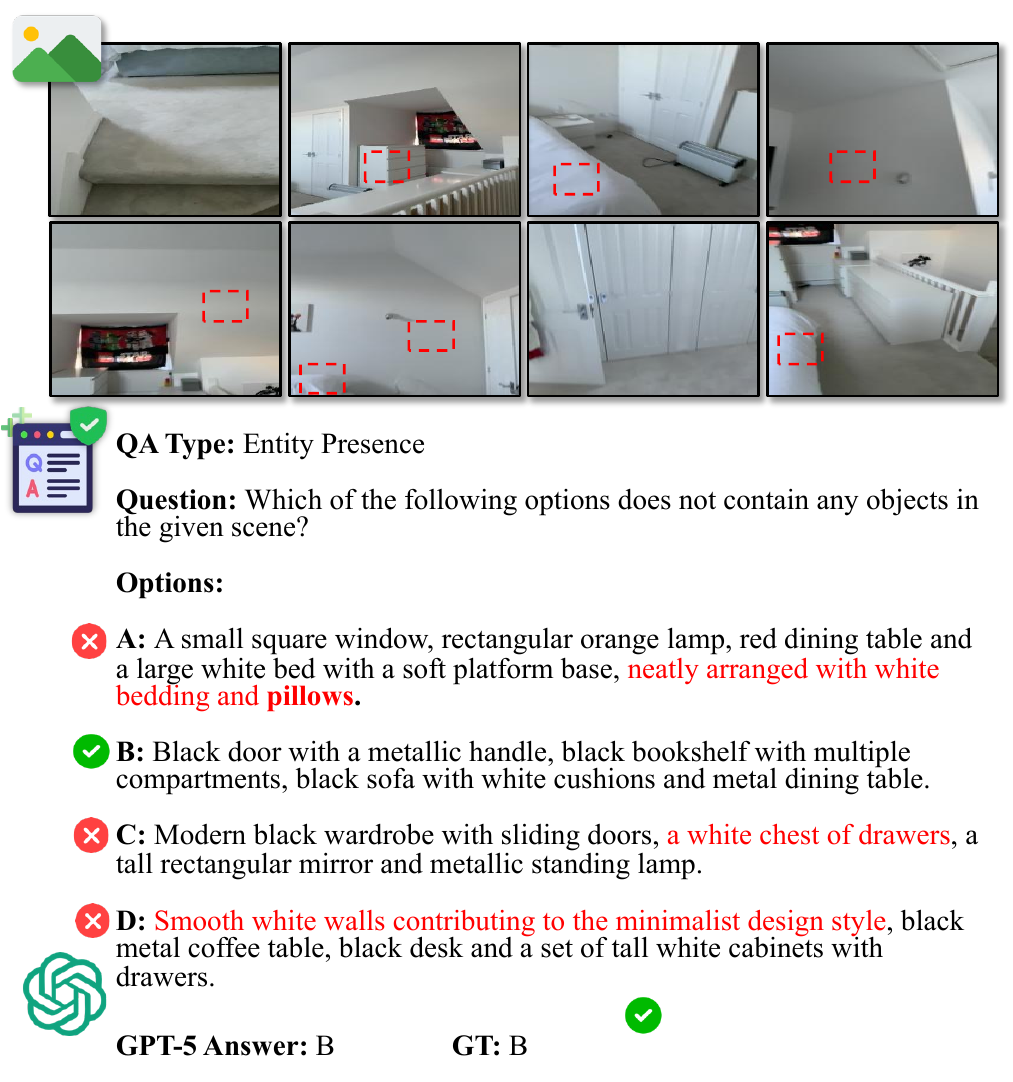}
    \caption{Visualization of Entity Presence}
    \label{ep2}
\end{figure}

\begin{figure}
    \centering
    \includegraphics[width=0.6\linewidth]{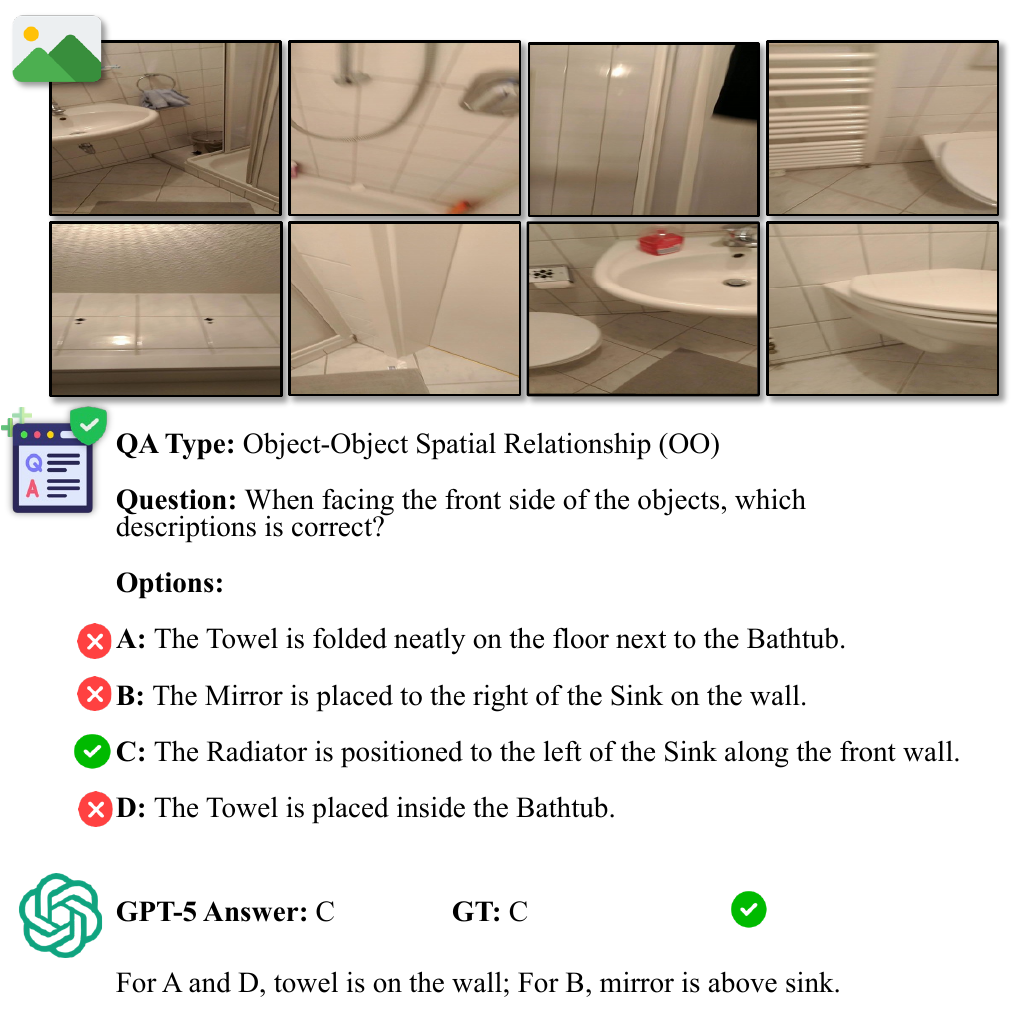}
    \caption{Visualization of Object-Object Spatial Relationship}
    \label{oo1}
\end{figure}

\begin{figure}
    \centering
    \includegraphics[width=0.6\linewidth]{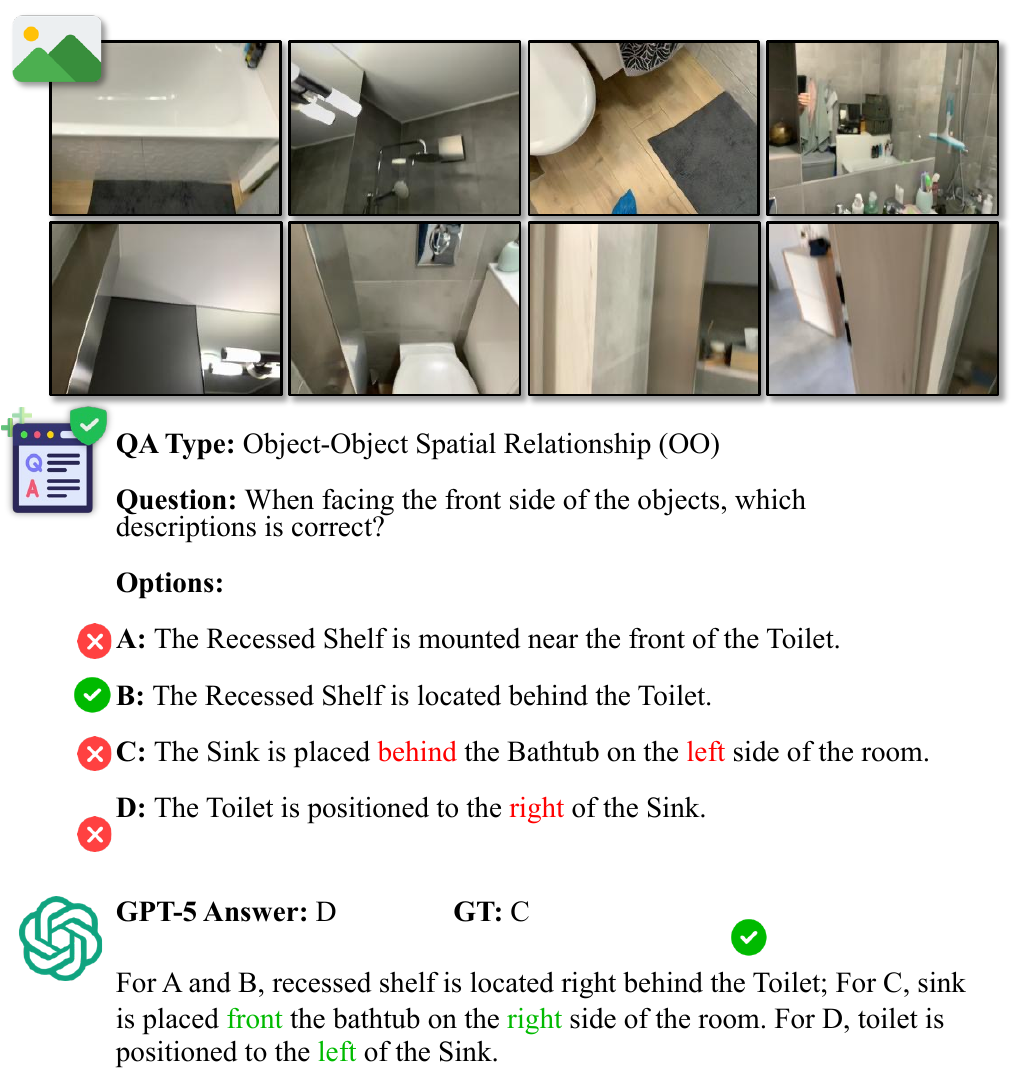}
    \caption{Visualization of Object-Object Spatial Relationship}
    \label{oo2}
\end{figure}

\begin{figure}
    \centering
    \includegraphics[width=0.6\linewidth]{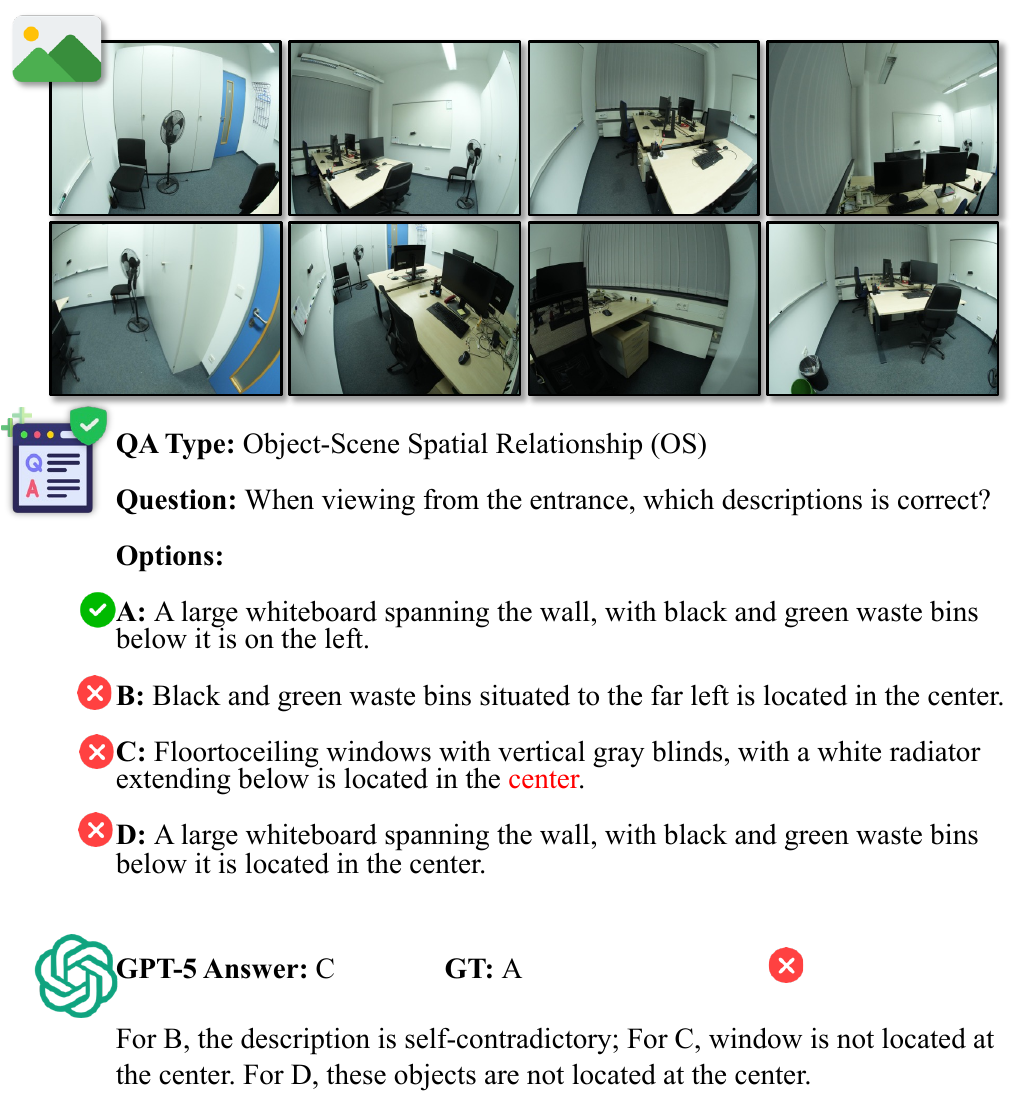}
    \caption{Visualization of Object-Scene Spatial Relationship}
    \label{os1}
\end{figure}

\begin{figure}
    \centering
    \includegraphics[width=0.6\linewidth]{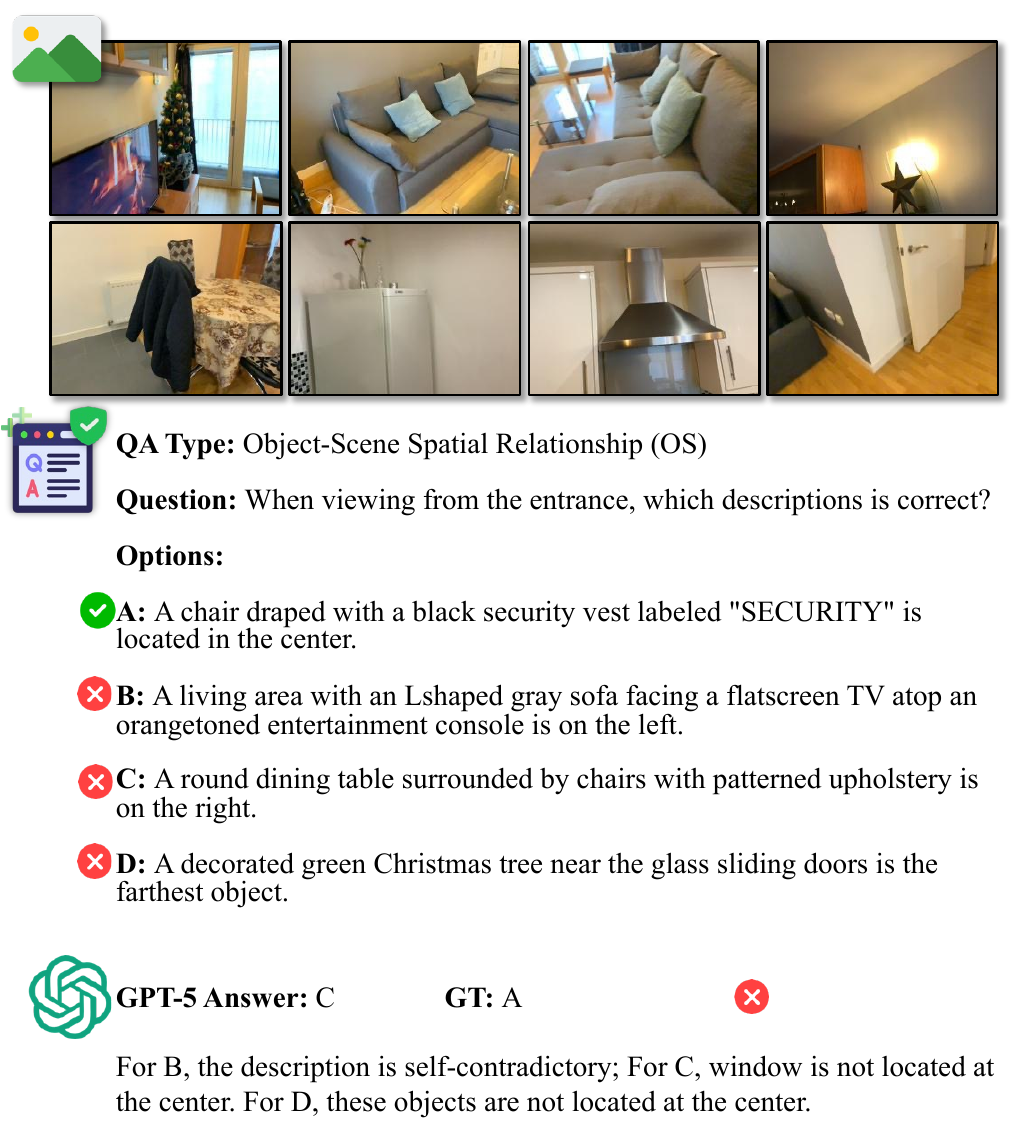}
    \caption{Visualization of Object-Scene Spatial Relationship}
    \label{os2}
\end{figure}

\begin{figure}
    \centering
    \includegraphics[width=0.6\linewidth]{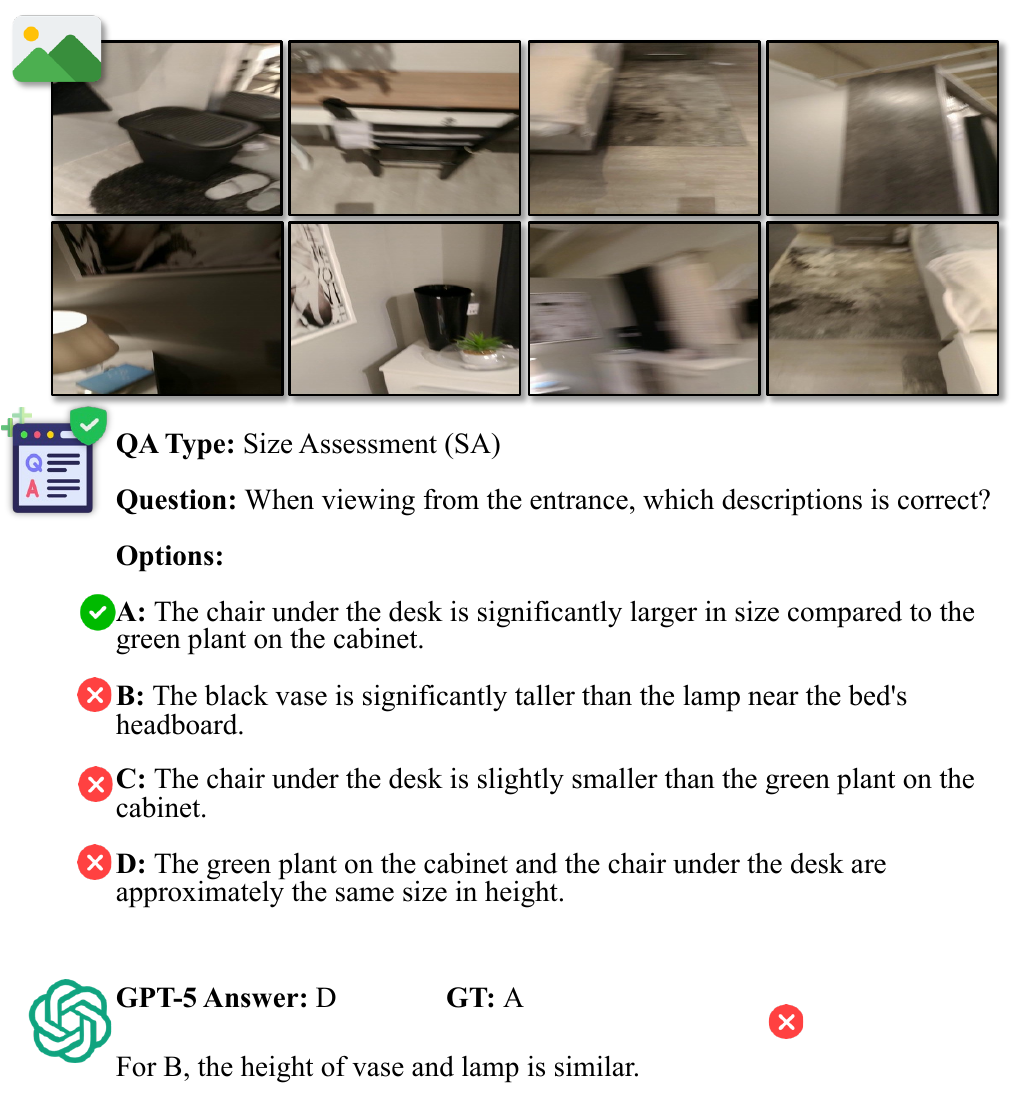}
    \caption{Visualization of Size Assessment}
    \label{sa1}
\end{figure}

\begin{figure}
    \centering
    \includegraphics[width=0.6\linewidth]{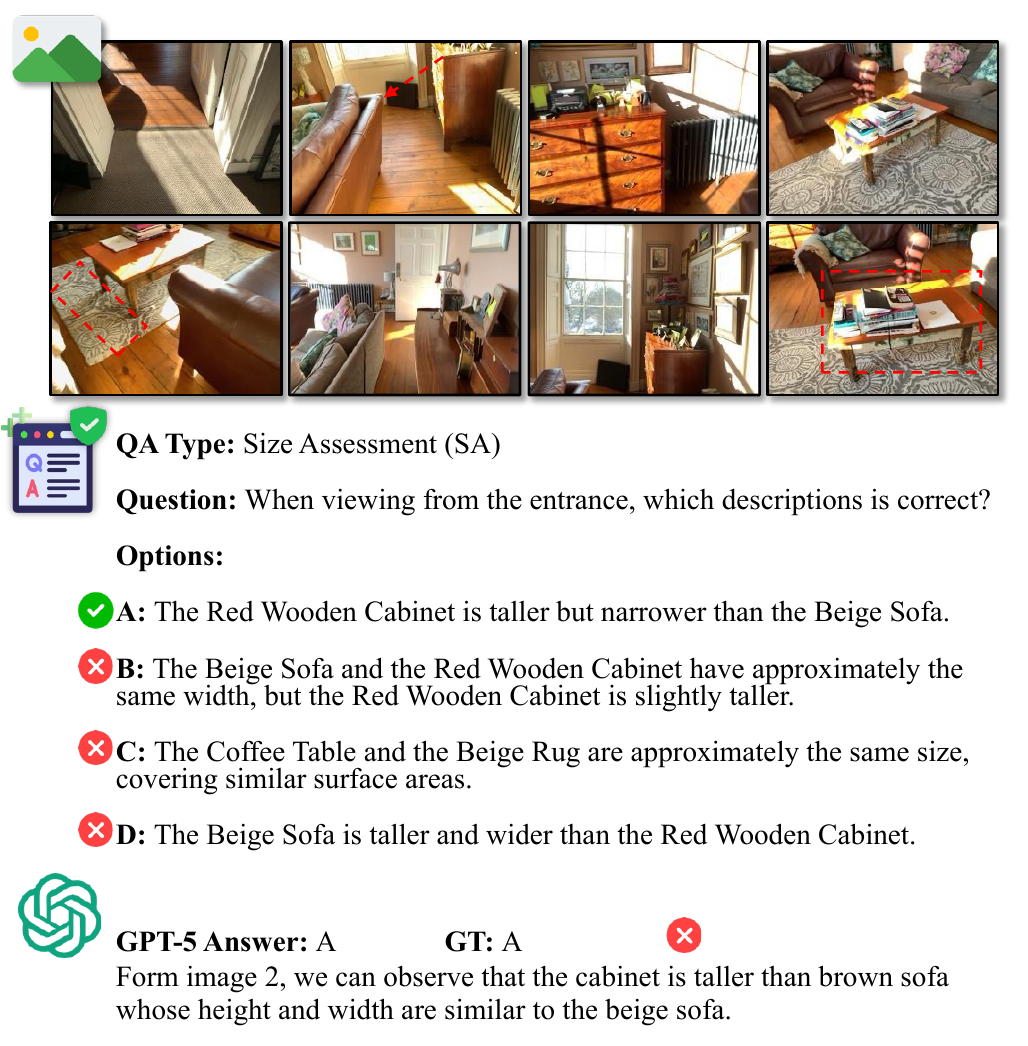}
    \caption{Visualization of Size Assessment}
    \label{sa2}
\end{figure}

\begin{figure}
    \centering
    \includegraphics[width=0.6\linewidth]{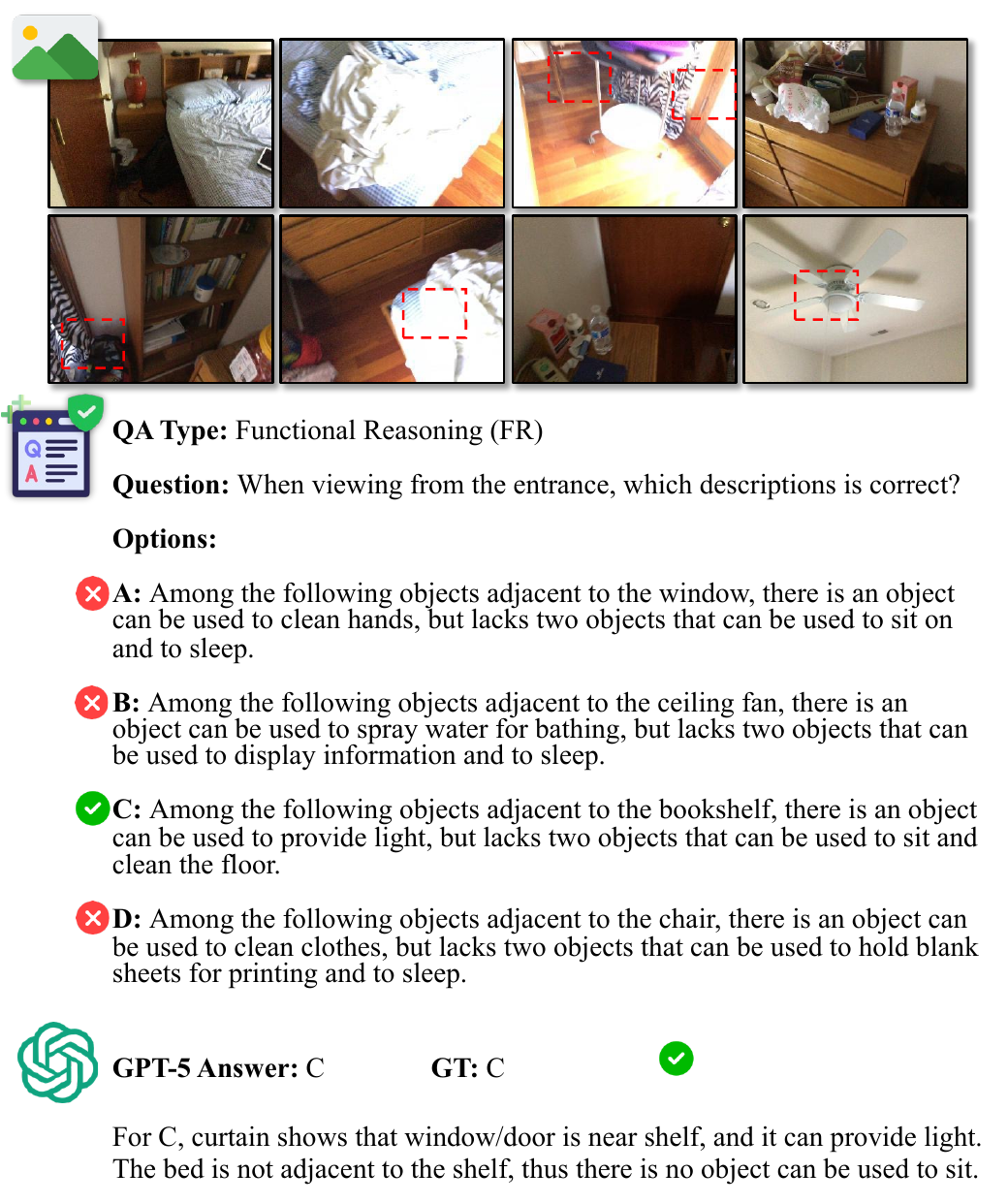}
    \caption{Visualization of Functional Reasoning}
    \label{fr1}
\end{figure}

\begin{figure}
    \centering
    \includegraphics[width=0.6\linewidth]{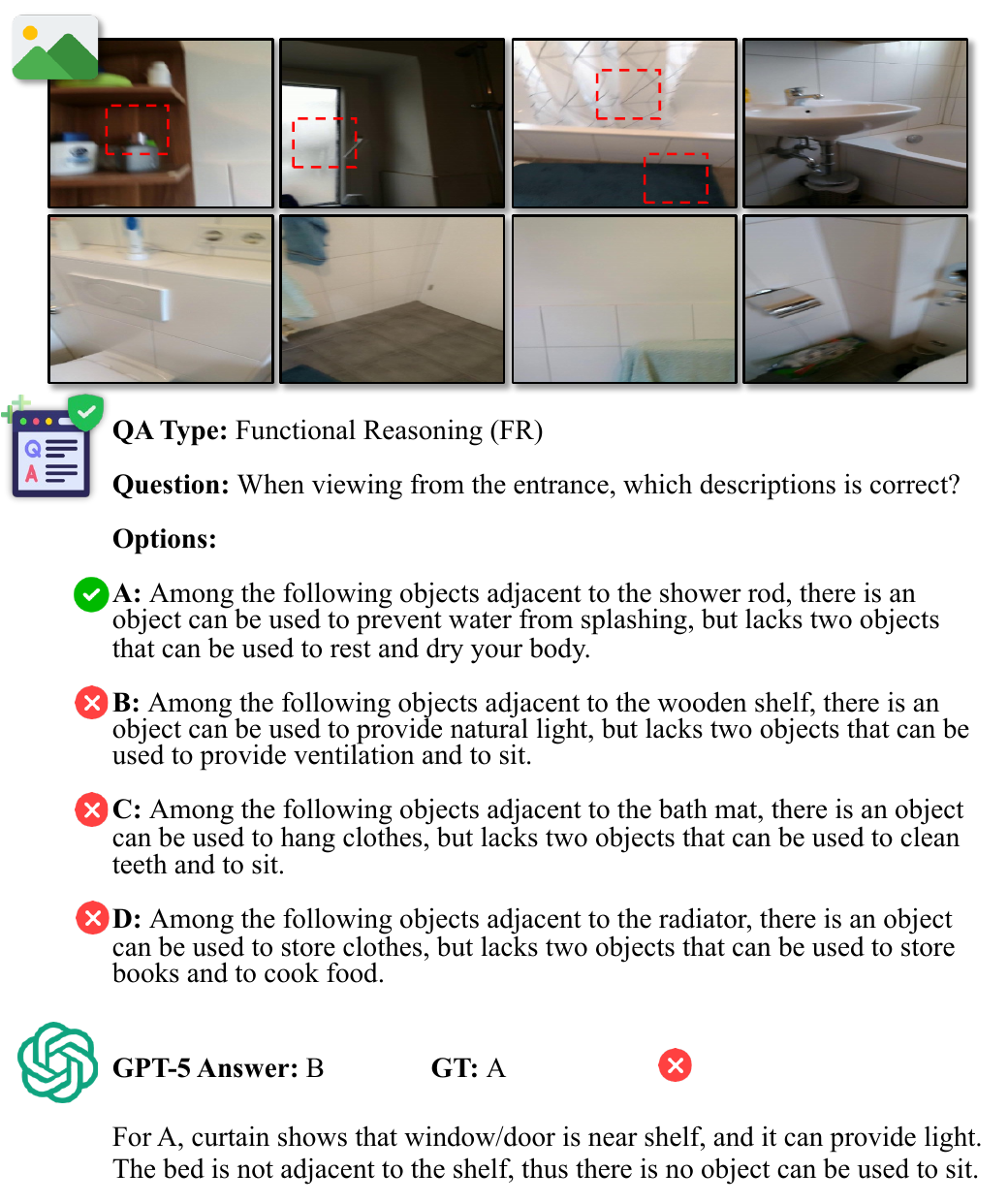}
    \caption{Visualization of Functional Reasoning}
    \label{fr2}
\end{figure}

\begin{figure}
    \centering
    \includegraphics[width=0.6\linewidth]{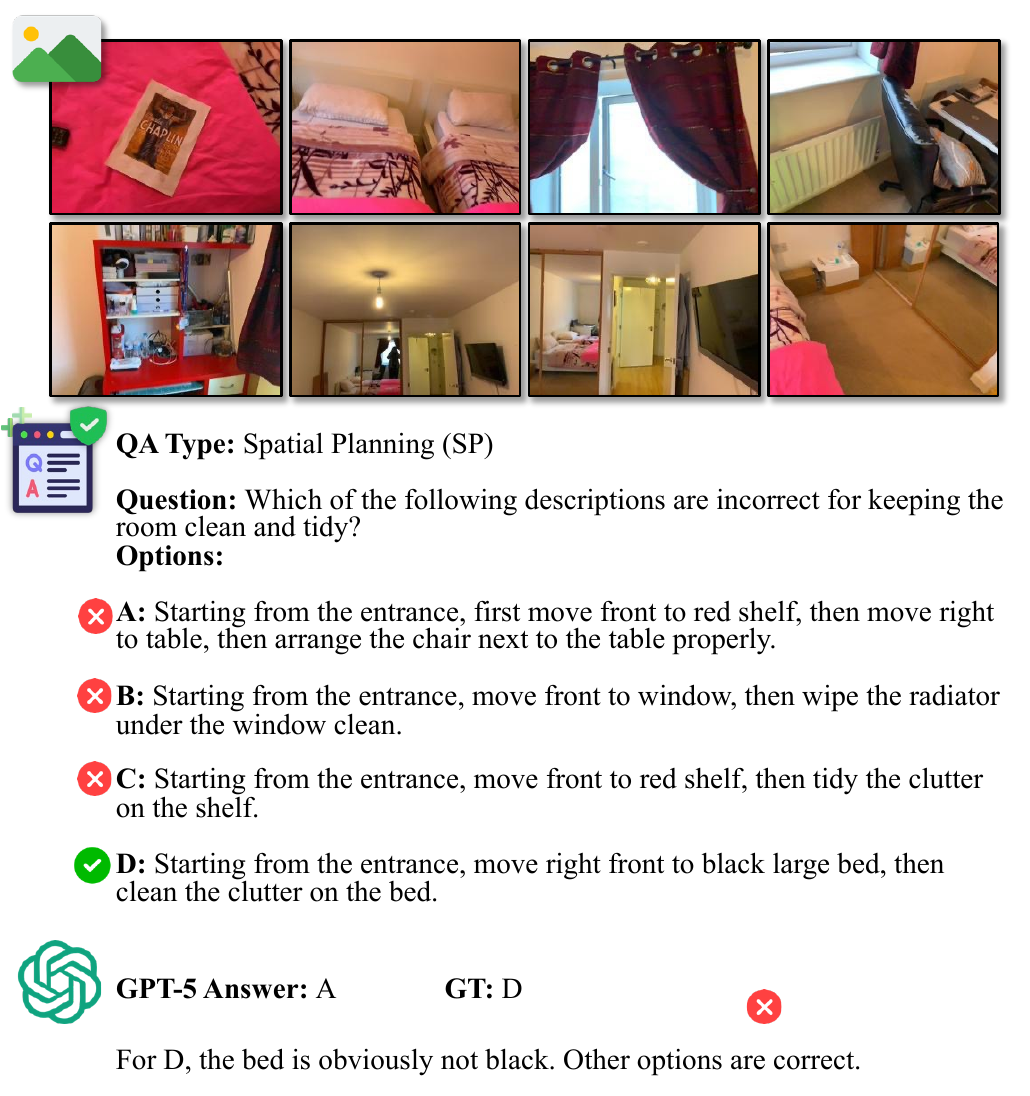}
    \caption{Visualization of Spatial Planning}
    \label{sp1}
\end{figure}

\begin{figure}
    \centering
    \includegraphics[width=0.6\linewidth]{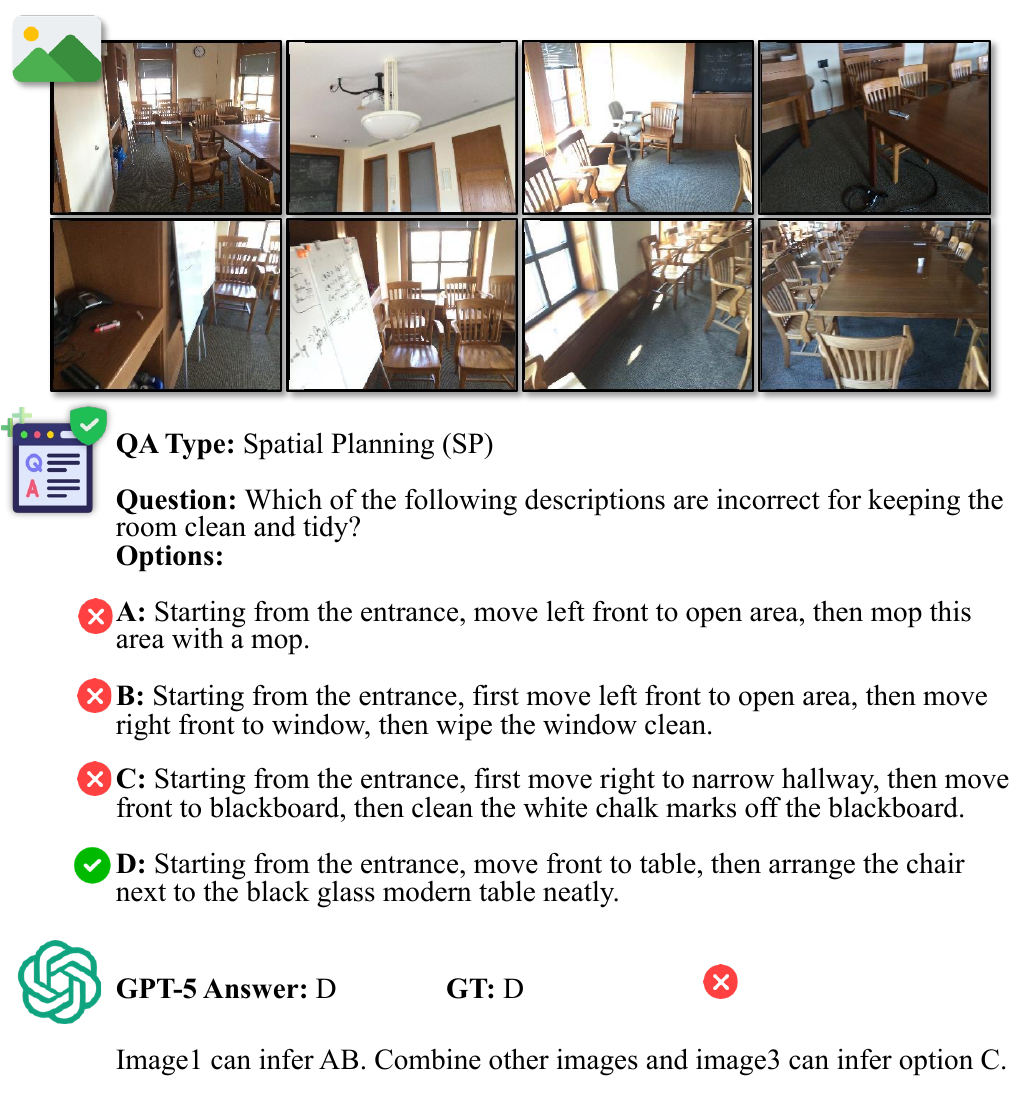}
    \caption{Visualization of Spatial Planning}
    \label{sp2}
\end{figure}

\begin{figure}
    \centering
    \includegraphics[width=0.6\linewidth]{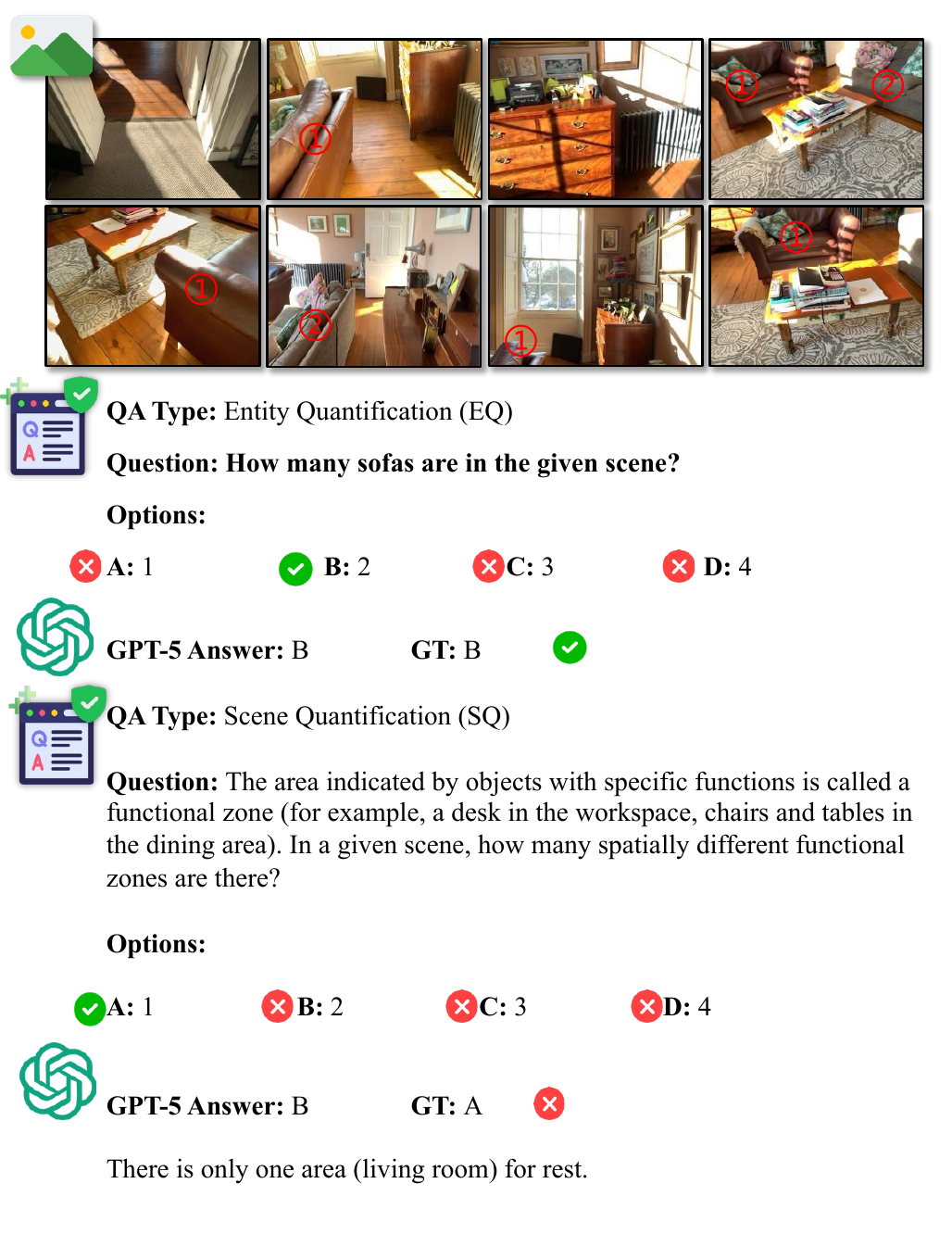}
    \caption{Visualization of Entity Quantification and Scene Quantification}
    \label{eq sq 1}
\end{figure}

\begin{figure}
    \centering
    \includegraphics[width=0.6\linewidth]{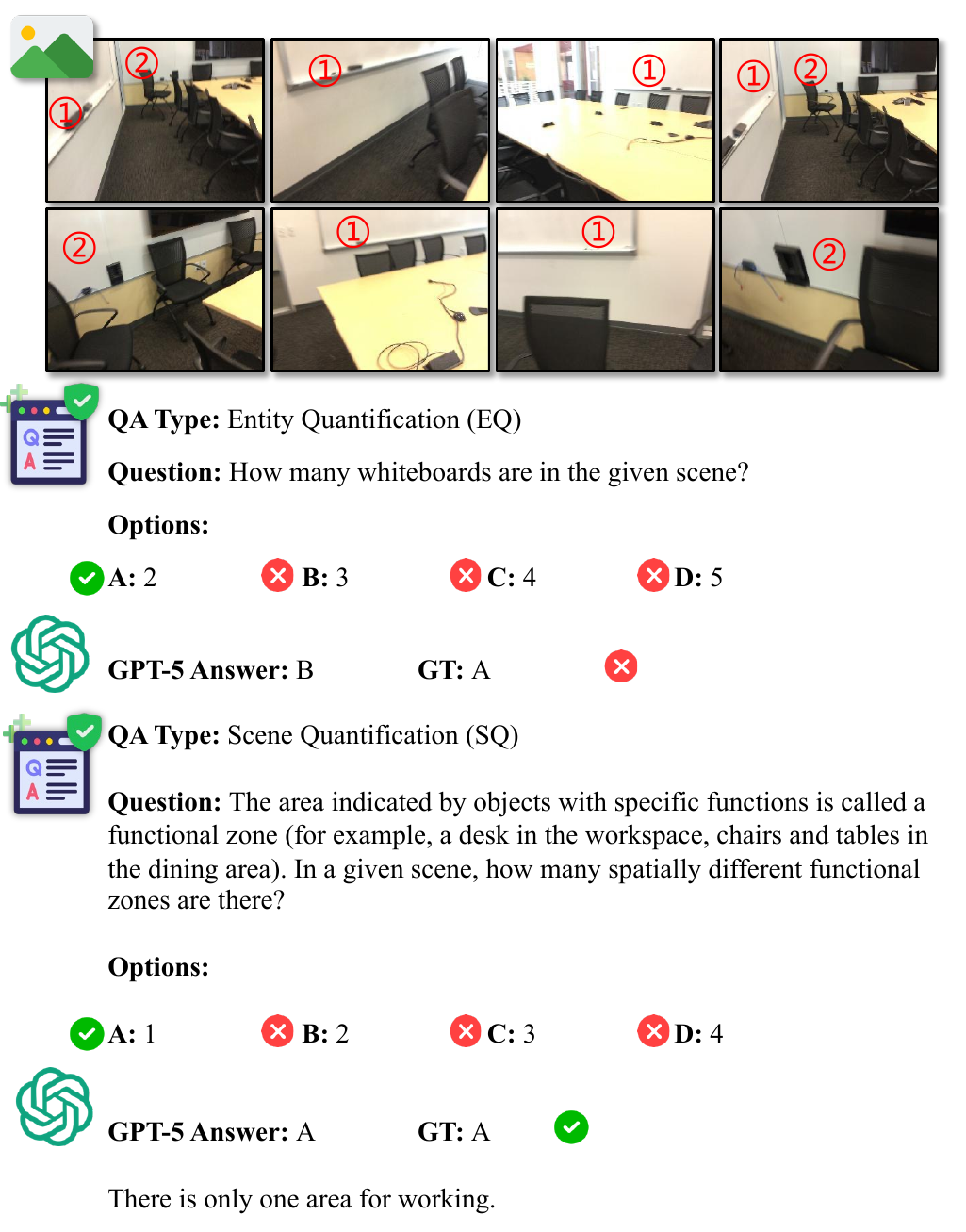}
    \caption{Visualization of Entity Quantification and Scene Quantification}
    \label{eq sq 2}
\end{figure}

We demonstrate more QA cases in this section. Specifically, figures of EP are shown in Fig.~\ref{ep1} and~\ref{ep2}; figures of OO are shown in Fig.~\ref{oo1} and ~\ref{oo2}; figures of OS are shown in Fig.~\ref{os1} and ~\ref{os2}; figures of SA are shown in Fig.~\ref{sa1} and ~\ref{sa2}; figures of FR are shown in Fig.~\ref{fr1} and ~\ref{fr2}; figures of SP are shown in Fig.~\ref{sp1} and ~\ref{sp2}; figures of EQ and SQ are shown in Fig.~\ref{eq sq 1} and ~\ref{eq sq 2}.
\label{visualization QA}


\end{document}